\documentclass[11pt]{article}

\usepackage[table]{xcolor}

\usepackage[preprint]{acl}

\usepackage{amsmath}
\usepackage{amssymb}
\usepackage{bm}

\usepackage{times}
\usepackage{latexsym}
\usepackage[T1]{fontenc}
\usepackage[utf8]{inputenc}
\usepackage{microtype}
\usepackage{inconsolata}

\usepackage{graphicx}
\usepackage{multirow}
\usepackage{array}
\usepackage{booktabs}

\definecolor{bannerLight}{HTML}{F2F2F2}
\definecolor{bannerGray}{HTML}{E7E6E6}
\definecolor{dasmarkGold}{HTML}{FCF1CC}

\newcommand{\BGL}{\cellcolor{bannerLight}}
\newcommand{\BGG}{\cellcolor{bannerGray}}
\newcommand{\DP}{\cellcolor{dasmarkGold}\bfseries}

\definecolor{hdrGray}{HTML}{E7E6E6}
\definecolor{dasGold}{HTML}{FCF1CC}

\title{Is This Your Agent? Identity Watermarking
for Agent Systems \\in Dynamic Environments}

\author{
  \textbf{Yuwei Zhong}\textsuperscript{1}%
  \thanks{Equal contribution.}\quad
  \textbf{Yu Cheng}\textsuperscript{1,2}\footnotemark[1]\quad
  \textbf{Jiuan Zhou}\textsuperscript{1,2}\quad
  \textbf{Weizhi Liu}\textsuperscript{1}\\
  \textbf{Yi Xie}\textsuperscript{1}\quad
  \textbf{Yongkang Hu}\textsuperscript{1}\quad
  \textbf{Yuan Xie}\textsuperscript{1,2}\quad
  \textbf{Zhaoxia Yin}\textsuperscript{1}%
  \thanks{Corresponding author.}\\
  \textsuperscript{1}East China Normal University\\
  \textsuperscript{2}Shanghai Innovation Institute
}

\begin{document}
\maketitle
\begin{abstract}
Agent systems deployed in dynamic environments continuously update
their memories, skills, and runtime configurations, leading to
evolving trajectories and challenges for persistent identity
authentication. Existing text and agent watermarking methods
struggle to achieve stable watermark embedding under dynamic
environments, while embedded watermarks may become difficult to
recover due to environmental changes.
This paper proposes \textbf{DASMark}, an identity watermarking
framework for dynamic agent systems. DASMark introduces
\textbf{Action-Preserving Embedding (APE)}, which learns semantically
similar but distinguishable decision-expression channels from diverse
interaction trajectories to enable stable watermark embedding in
dynamic systems. It further introduces
\textbf{Shift-Robust Recovery (SRR)} to mitigate context-induced
shifts and improve watermark recovery stability under dynamic
environments.
In dynamic settings, DASMark achieves 98.5\% bit-decoding accuracy and maintains 95.2\% accuracy under attack scenarios, while preserving task utility comparable to unwatermarked agents.
\end{abstract}

\section{Introduction}

With the rapid development of LLM-driven agent systems, dynamic
agents that can perceive environments, invoke tools, and update
their states are becoming an important paradigm for next-generation
autonomous systems \citep{yao2023react,liu2024agentbench}. Dynamic
agents continuously update their memories, accumulate reusable
skills, and adapt their behavior based on task feedback
\citep{shinn2023reflexion,wang2024tmlr-voyager,
xu2025amem,cheng2026tame,park2023generative},
resulting in evolving behavior processes and runtime contexts.
However, such evolution increases the difficulty of behavior
attribution, making risks such as identity impersonation,
unauthorized invocation, and proprietary agent replication more
prominent \citep{deng2025agentsunderthreat,triedman2025multiagent,
huang2026agentmark}. 
To address these challenges, \textbf{agent watermarking} has emerged as a promising approach for identity authentication and behavior attribution by embedding verifiable signals into generated content, planning behaviors, or execution trajectories.

\begin{figure}[t]

  \centering
  \includegraphics[width=\columnwidth]{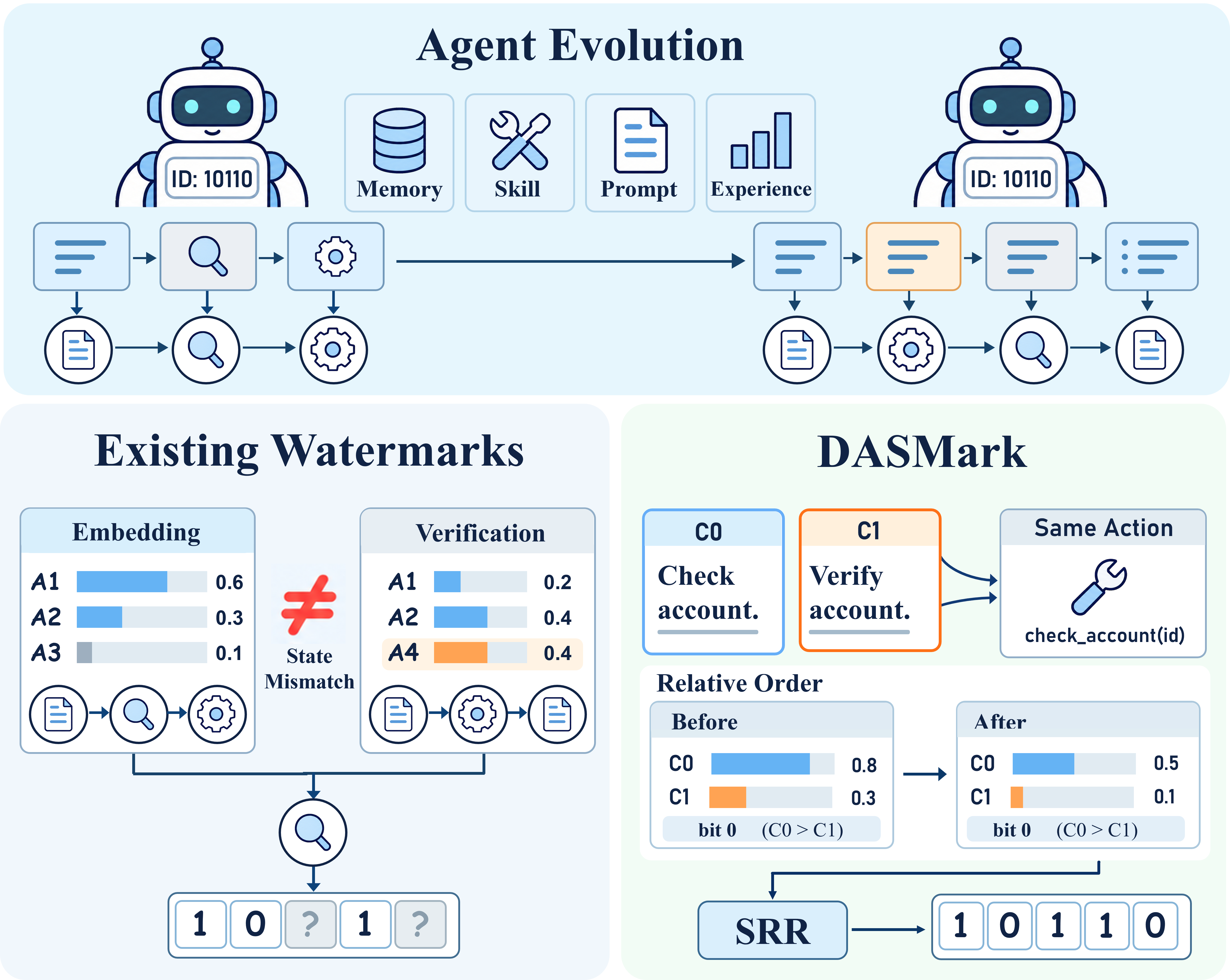}
\caption{
Identity watermark recovery under dynamic agent evolution.
Dynamic updates to memory, skills, prompts, and experience shift
decision-expression distributions and challenge watermark recovery.
DASMark preserves tool actions during embedding and uses SRR with
relative score comparison to achieve stable recovery from final
trajectories.
}
  \label{fig:motivation}
\end{figure}

Existing agent watermarking methods face two key
challenges in dynamic environments.
First, \textbf{embedding adaptability}.
Text watermarking and some behavioral watermarking
methods embed signals through token sampling or
action selection
\citep{kirchenbauer2023watermark,dathathri2024scalable,
huang2026agentmark}.
As agent systems continually update their memories,
skills, and execution strategies, changes in generation
contexts and available actions can affect the consistency
of watermark embedding.
Second, \textbf{recovery robustness}.
Behavioral watermarking methods recover identities
from action distributions or trajectory structures,
and some require additional probability information
for decoding
\citep{huang2026agentmark,an2026seqwm}.
Continual system updates can shift the distributions
of watermark-bearing signals, reducing recovery
stability.
Adversarial modifications, such as rewriting decision
expressions or deleting and reordering trajectory
segments, may further weaken these signals or disrupt
the structures required for decoding.
Dynamic agent watermarking therefore requires reliable
identity embedding across continual system updates
and robust recovery under both environmental changes
and adversarial perturbations.

To tackle these challenges, this paper proposes \textbf{DASMark (Dynamic Agent Systems Watermark)}, a learning-based identity watermarking framework for dynamic agent systems. As shown in Figure~\ref{fig:motivation}, DASMark focuses on two key aspects of dynamic environments: \textbf{embedding adaptability} and \textbf{recovery stability}. DASMark introduces a \textbf{behavior-preserving embedding mechanism} that learns stable decision expression channels and embeds identity information into natural decision expressions, achieving stable embedding. Furthermore, DASMark proposes \textbf{Shift-Robust Recovery (SRR)}, which recovers identity bits using relative decoding scores between adjacent covers, improving recovery stability under dynamic environments. Experimental results show that DASMark achieves a 5.1\% higher average bit decoding accuracy than SOTA methods under various attacks, while maintaining task utility and avoiding additional trajectory overhead. These results demonstrate the effectiveness of DASMark for identity attribution in dynamically deployed agent systems. The main contributions are summarized as follows:
\begin{itemize}
    \item \textbf{A watermarking framework for dynamic agent systems.}
    DASMark targets identity attribution challenges in dynamic environments by embedding multi-bit identity information into sequences of decision expressions, enabling agent identity verification under dynamic deployment.

    \item \textbf{Action-preserving embedding and shift-robust recovery mechanisms.}
    Action-Preserving Embedding (APE) learns stable decision-expression channels for reliable identity embedding, while Shift-Robust Recovery (SRR) recovers identity bits through relative decoding scores between adjacent covers, mitigating recovery instability caused by dynamic environments.

   \item \textbf{Experiments cover cross-domain tasks, dynamic environments,
and various attack scenarios.} DASMark achieves a 5.1\% higher average
bit-decoding accuracy than SOTA methods under attack settings, while
maintaining task utility and trajectory length comparable to those of
unwatermarked agents.

\end{itemize}
\section{Related Work}
\label{sec:related-work}

\subsection{LLM Text Watermarking}

LLM text watermarking embeds signals
into generated text for content attribution and verification.
Early studies employ vocabulary partitioning and
sampling biases for watermark embedding
\citep{aaronson2023watermark,kirchenbauer2023watermark}.
Subsequent research explores generation quality,
detection reliability, and information capacity,
including distortion-free watermarking
\citep{kuditipudi2023robust},
scalable watermarking \citep{dathathri2024scalable},
unbiased watermarking \citep{hu2024iclr-unbiased},
and multi-bit watermarking
\citep{yoo-etal-2024-advancing}.
These methods primarily address attribution of generated text, whereas we study identity recovery from agent trajectories under dynamically changing runtime contexts.

\subsection{Agent Watermarking}

Agent watermarking extends identity embedding
to planning, behaviors, and long-term states.
Agent Guide \citep{huang2025agentguide},
AgentMark \citep{huang2026agentmark},
and SeqWM \citep{an2026seqwm}
investigate behavioral watermarking through behavioral biases,
distribution-preserving sampling, and historical
behavior correlations, respectively.
ActHook \citep{meng2026acthook},
TRACE \citep{gao2026trace},
and MemMark \citep{zhang2026memmark}
further extend watermarking to execution trajectories
and memory states.
Existing approaches primarily rely on action selection,
behavioral patterns, or system states, leaving naturally
generated decision expressions largely unexplored.
The reliability of watermark recovery under evolving
runtime contexts also remains an open question.

\subsection{Dynamic Agent Systems}

LLM agent systems increasingly operate in dynamic
environments, continually updating their memory,
skills, and execution strategies through interaction.
Generative Agents \citep{park2023generative} and
MemGPT \citep{packer2023memgpt} investigate memory
and context management, while A-MEM \citep{xu2025amem}
and TAME \citep{cheng2026tame} explore adaptive memory
organization and continual memory evolution.
Beyond memory, Evo-Harness \citep{wei2026evoharness}
and SafeCoEvo \citep{cheng2026safecoevo} investigate
the evolution of execution configurations and safety
mechanisms.
These continual updates reshape agents' decision
contexts and interaction trajectories.
Whereas these studies focus on agent adaptation,
DASMark studies identity watermarking for continually
updated agent systems in dynamic environments.

\section{Method}
\label{sec:method}

\subsection{Overview}
\label{sec:overview}

As tasks progress, agent systems continually update their
memory, skills, prompts, and execution configurations,
reshaping their decision contexts and trajectory expressions.
Such changes can alter the expression distributions that
carry watermark signals, making identities embedded across
successive interactions harder to recover consistently.
The challenge is to maintain reliable identity signals as
the agent system changes, while preserving its task execution.
We propose DASMark, a learned identity watermarking framework
that embeds signals in agent decision expressions without
modifying tool actions and supports reliable identity recovery
from trajectories in dynamic environments.

In dynamic agent systems, the state at decision step $t$ is denoted as $s_t=(h_t,e_t)$, where $h_t$ represents the interaction history and $e_t$ represents the dynamic context determined by memory, skills, and runtime configurations. As the trajectory accumulates, the runtime context continuously evolves as $e_{t+1}=\mathcal U(e_t,\tau_{\leq t})$, causing the same behavior to potentially produce different decision expression distributions. Consequently, watermarking methods based on fixed distribution assumptions may fail to maintain long-term stability.

\begin{figure*}[t]
    \centering
    \includegraphics[width=0.9\textwidth]{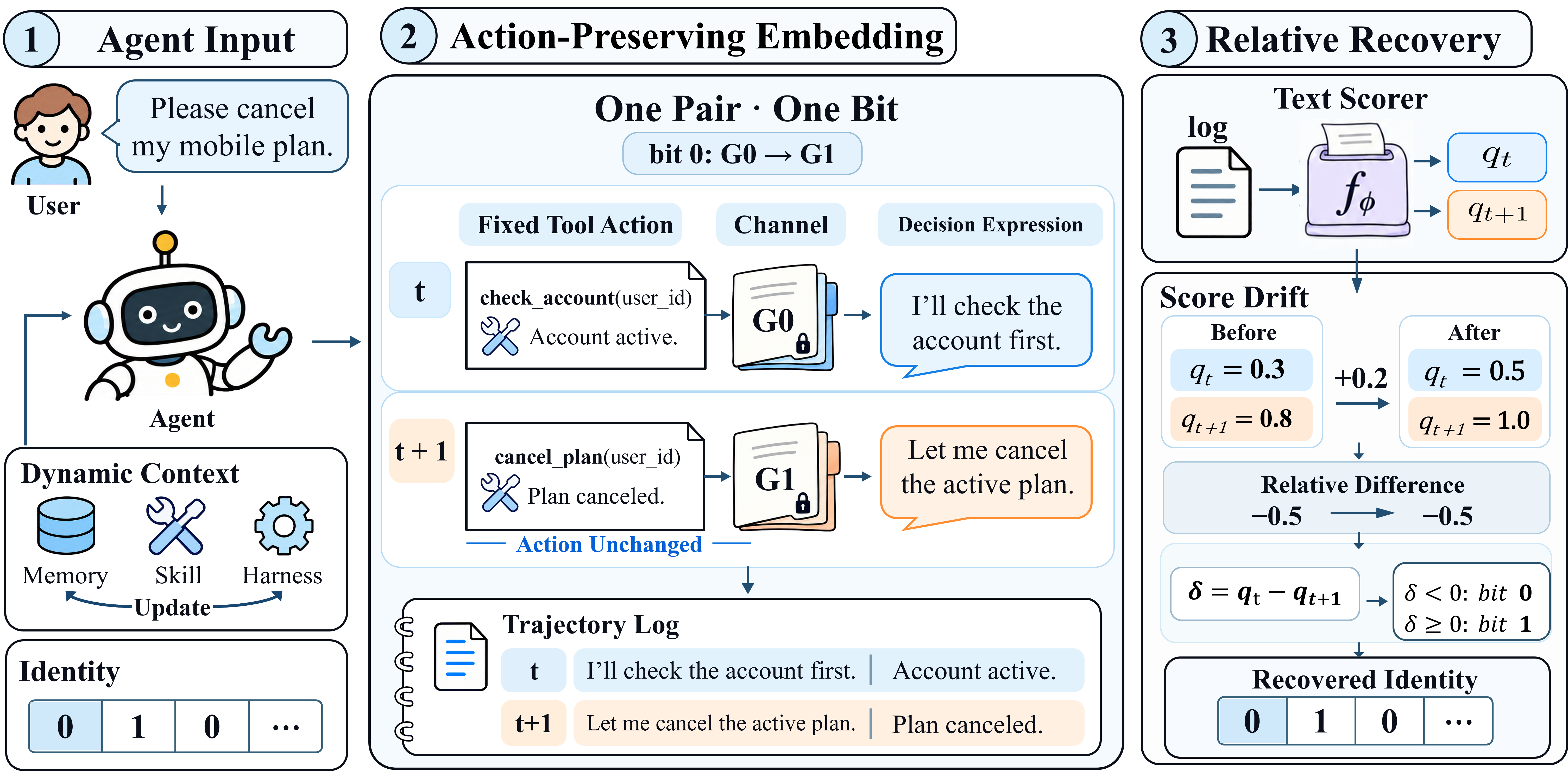}
    \caption{
    Overview of DASMark.
    DASMark consists of three stages: watermark signal construction,
    action-preserving embedding, and shift-robust recovery.
    In the offline stage, DASMark discovers stable decision-expression
    variations from real interaction states, learns two decision-expression
    channels, and trains a receiver using generated covers.
    During deployment, identity bits are embedded through channel ordering
    without changing tool actions, and recovered through relative score
    comparison between adjacent covers under dynamic context shifts.
    }
    \label{fig:framework}
\end{figure*}

As shown in Figure~\ref{fig:framework}, the overall pipeline
of DASMark consists of offline learning, online embedding,
and identity recovery.
DASMark first learns watermark generators and a decoder
from decision expressions across diverse agent states,
then embeds identity bits through paired expressions
during task execution and recovers them from the
resulting trajectory.
Two key mechanisms address the challenges posed by
dynamic environments.
First, the behavior-preserving embedding mechanism learns
stable expression differences across diverse runtime states,
enabling identity information to adapt to dynamic changes.
Second, the shift-robust recovery mechanism exploits relative
decoding relationships between adjacent expressions to
improve identity recovery stability during long-term operation.


\subsection{Watermark Signal Construction}
\label{sec:signal-construction}

The key challenge in constructing a watermark signal is to discover stable
decision-expression variations that are independent of task semantics from
task execution trajectories. Partitioning texts from different tasks into two channels
risks conflating watermark signals with task-specific
features, such as tool types, task categories, and
execution outcomes. DASMark instead extracts interaction
states preceding tool calls from the base agent's
execution trajectories and learns variations in decision
expressions while keeping tool actions fixed.

Given the sampling state $s_i$ and the tool action $a_i$ determined by the base agent, the candidate generation stage keeps the action and its parameters unchanged while resampling only the corresponding natural language decision expressions. This process constructs candidate expressions with the same underlying behavior:
\begin{equation}
\mathcal{Y}_i=\{y_{ij}\}_{j=1}^{K},
\qquad
y_{ij}\sim p_\theta(\cdot|s_i,a_i).
\label{eq:candidate-generation}
\end{equation}

DASMark then maps candidate expressions into a normalized semantic space using
a frozen semantic encoder $E(\cdot)$ \citep{reimers-gurevych-2019-sentence}:
\begin{equation}
z_{ij}=\frac{E(y_{ij})}{\|E(y_{ij})\|_2}.
\label{eq:semantic-representation}
\end{equation}

Here, $p_\theta$ denotes the unwatermarked base language model,
and $E(\cdot)$ is the frozen semantic encoder.
Candidate expressions are encoded and normalized to obtain
semantic representations $z_{ij}$, with implementation details
provided in Appendix~\ref{app:watermark-signal}.

DASMark subsequently aggregates candidates within each state according to
semantic similarity and constrains each cluster to maintain semantic
consistency, length similarity, and expression diversity. Since clustering is
performed only within the same state, the learned expression directions are
less affected by task semantics and text length variations.
However, representations from different states may still be influenced by user
requests, historical contexts, and tool semantics. To remove state-related
variations, DASMark first centers candidate representations within each state
and identifies the subspace dominated by task and state changes through
singular value decomposition. The representations are then projected onto the
orthogonal complement of this subspace:
\begin{equation}
r_{ij}=(I-V_rV_r^\top)(z_{ij}-\mu_i),
\label{eq:residual-representation}
\end{equation}

where $\mu_i$ denotes the candidate representation center of the $i$-th state,
and $V_r$ represents the subspace composed of dominant task and state variation
directions.

Based on the residual representations, DASMark further identifies a
cross-state shared expression variation direction. To prevent states with more
candidates from dominating the optimization, DASMark assigns equal weights to
different states and maximizes the within-state expression variation:
\begin{equation}
v^*
=
\underset{\|v\|_2=1}{\arg\max}
\sum_{i=1}^{N}
\frac{1}{|\mathcal C_i|}
\sum_{j\in\mathcal C_i}
(r_{ij}^{\top}v)^2 .
\label{eq:expression-direction}
\end{equation}

The obtained direction $v^*$ represents a shared expression variation axis supported by multiple states. DASMark ranks candidates according to their projection scores along this direction and selects the two highest-scoring and two lowest-scoring candidates within each state as supervision targets for two decision expression channels. Candidates are discarded if they are inconsistent with the current history or predefined action, or contain factual errors, parameter deviations, or premature assertions of tool results. The resulting supervised dataset for sender training is defined as $\mathcal{D}_{\mathrm{send}}=\{(s_i,a_i,y,b)\}$, where $b\in\{0,1\}$ denotes the expression channel label.

\begin{figure*}[t]
    \centering
    \includegraphics[width=0.9\linewidth]{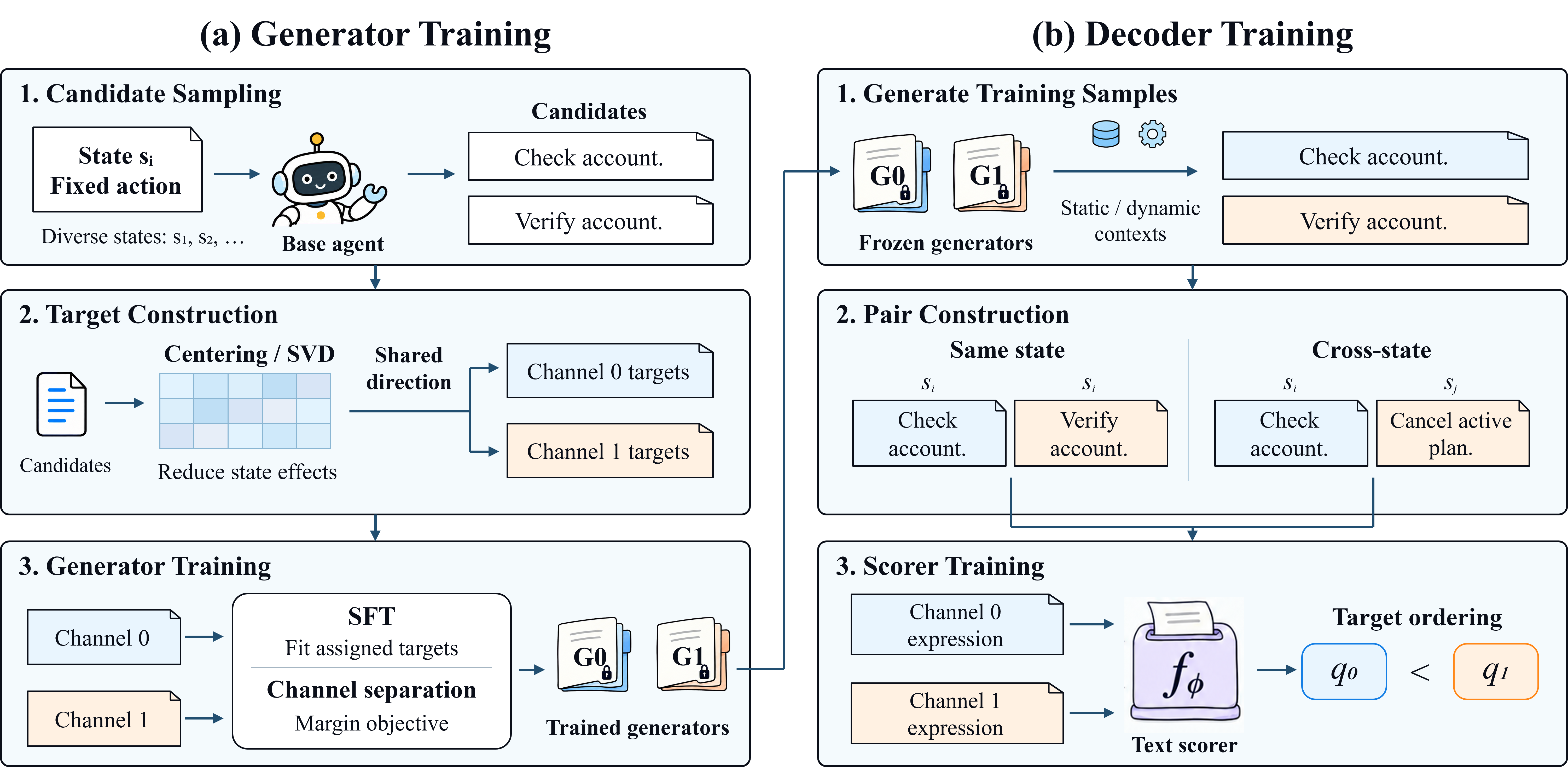}
\caption{
Training process of DASMark.
DASMark first learns two distinguishable decision-expression
channels under the same state and fixed action, and then trains
the decoder to recover watermark bits through relative score
comparison.
}
    \label{fig:embedding}
\end{figure*}

\subsection{Action-Preserving Embedding}

After watermark signal construction, DASMark transforms
the identified expression variations into controllable
generation capabilities, enabling identity information
to be embedded through distinct decision-expression
channels. Specifically, DASMark selects candidates with
high and low projection scores along the identified
expression direction within each state, using them as
supervision targets for two channels to construct the
sender training dataset $\mathcal{D}_{\mathrm{send}}$
(see Appendix~\ref{app:watermark-signal} for details).
Two independent LoRA branches, $G_0$ and $G_1$, are
then trained to learn distinguishable expression patterns.

As shown in Figure~\ref{fig:embedding}, both branches
are conditioned on the same complete state $s_i$ and
predetermined tool action $a_i$, generating only the
natural language decision expression corresponding to
that action. Since the tool name and invocation arguments
are determined before expression generation, channel
selection affects only the decision expression without
altering the tool action, satisfying the action-preserving
constraint $a_t^{\mathrm{wm}}=a_t^{\mathrm{base}}$.
To minimize interference with the original capabilities
of the base model, DASMark freezes its parameters and
optimizes only the two lightweight LoRA branches,
allowing them to learn channel-specific expression
patterns while sharing the underlying language capabilities.

However, supervised fine-tuning (SFT) alone may cause
the two branches to learn similar generation distributions,
limiting their distinguishability. To address this issue,
DASMark introduces an inter-channel margin constraint
alongside the SFT objective. This encourages each target
expression to have not only a high likelihood under its
corresponding channel but also a relatively lower
likelihood under the alternative channel.

Specifically, let $\ell_c$ denote the average token
log-likelihood of a target expression $y$ under branch
$G_c$, and define the channel margin as
$\Delta_b=\ell_b-\ell_{1-b}$. For a training sample
labeled with channel $b$, the sender is optimized using:

\begin{equation}
\mathcal{L}_{\mathrm{send}}
=
-\ell_b
+
\lambda\tau
\operatorname{softplus}
\left(
\frac{\gamma-\Delta_b}{\tau}
\right).
\end{equation}

The first term, $-\ell_b$, is the SFT loss, which
increases the likelihood of generating the target
expression under its corresponding channel.
The second term is the inter-channel margin constraint,
which improves channel distinguishability by penalizing
insufficient likelihood separation. When $\Delta_b$
falls below the target margin $\gamma$, a stronger
penalty encourages a higher relative likelihood under
the target channel. The parameters $\lambda$ and $\tau$
control the constraint weight and smoothness, respectively.

By jointly optimizing these objectives, APE improves
the separation between the two generation distributions
while learning the desired decision expressions.
Consequently, the two branches can produce
distinguishable expressions under identical state and
fixed-action conditions, providing the basis for
subsequent identity bit embedding and watermark recovery.

\subsection{Shift-Robust Recovery}
\label{sec:shift-robust-recovery}

Changes in dynamic agent environments may shift the distribution of decision expressions generated by the same watermark channel, causing systematic shifts in receiver scores and reducing the reliability of identity recovery. To mitigate such shifts, DASMark introduces Shift-Robust Recovery (SRR) to improve recovery stability under dynamic contexts.

After training the two decision expression channels, DASMark trains a receiver to identify the channel associated with each decision expression(see Appendix~\ref{app:training-details} for details), providing the basis for subsequent identity recovery.The receiver $f_\phi$ employs a byte-level bidirectional GRU \citep{cho-etal-2014-learning} to encode the complete decision expression and predicts the probability of belonging to channel 1, corresponding to $G_1$, as $p_i=\sigma(f_\phi(y_i))$. The receiver is trained on decision expressions generated by the frozen sender branches, reducing the distribution gap between training samples and actual watermark covers. For a batch of size $N$, the receiver is optimized using binary cross-entropy:

\begin{equation}
\mathcal{L}_{\mathrm{recv}}
=
\operatorname{BCE}(\mathbf{b},\mathbf{p}).
\end{equation}

During deployment, DASMark sequentially encodes identity bits using the channel ordering of two consecutive decision expressions. Specifically, $m_k=0$ is represented by the channel sequence $(0,1)$, whereas $m_k=1$ is represented by $(1,0)$. During verification, the original trajectory order is preserved, and each bit is recovered by comparing the receiver scores of two adjacent covers:

\begin{equation}
\widehat{m}_k
=
\mathbb{I}
\left[
f_\phi(y_{2k})-f_\phi(y_{2k+1})\geq 0
\right].
\end{equation}

The motivation for relative score comparison is that dynamic context changes may shift the decoder scores of the same channel, making fixed-threshold decisions unreliable. Since adjacent covers typically occur under similar local runtime contexts, DASMark assumes that their score shifts are correlated and can be approximated as a shared offset, i.e., $\varepsilon_{2k}\approx\varepsilon_{2k+1}$.

Let $\delta_k=f_\phi(y_{2k})-f_\phi(y_{2k+1})$ denote the original score difference. When the two covers experience score shifts $\varepsilon_{2k}$ and $\varepsilon_{2k+1}$, respectively, the shifted score difference becomes:

\begin{equation}
\delta_k'
=
\delta_k+
(\varepsilon_{2k}-\varepsilon_{2k+1})
\approx\delta_k.
\end{equation}

Thus, relative score comparison cancels the shared component of the shifts between adjacent covers, allowing identity recovery to rely primarily on their original channel-score ordering. As long as the residual shift does not reverse the sign of the score difference, the identity bit can be correctly recovered.


\section{Experiments}
\label{sec:experiments}
\begin{table*}[t]
\centering
\normalsize
\setlength{\tabcolsep}{2.8pt}
\renewcommand{\arraystretch}{1.0}
\resizebox{\textwidth}{!}{%
\begin{tabular}{@{} l p{1.3em} ccc p{1.3em} ccc p{1.3em} ccc @{}}
\toprule

\multicolumn{2}{c}{\multirow{2}{*}{\Large\textbf{Method}}}
  & \multicolumn{3}{c}{\large\textbf{Telecom}}
  & & \multicolumn{3}{c}{\large\textbf{CRMArena-Pro}}
  & & \multicolumn{3}{c}{\large\textbf{Retail}} \\

\cmidrule(lr){3-5}
\cmidrule(lr){7-9}
\cmidrule(lr){11-13}

& &
\textbf{Steps}\,$\boldsymbol{\downarrow}$ &
\textbf{SR}\,$\boldsymbol{\uparrow}$ &
\textbf{Bit Acc.}\,$\boldsymbol{\uparrow}$
& &
\textbf{Steps}\,$\boldsymbol{\downarrow}$ &
\textbf{SR}\,$\boldsymbol{\uparrow}$ &
\textbf{Bit Acc.}\,$\boldsymbol{\uparrow}$
& &
\textbf{Steps}\,$\boldsymbol{\downarrow}$ &
\textbf{SR}\,$\boldsymbol{\uparrow}$ &
\textbf{Bit Acc.}\,$\boldsymbol{\uparrow}$ \\

\midrule

\BGL{\large\bfseries\itshape No-Mem}
& \BGL & \BGL & \BGL & \BGL
& \BGL & \BGL & \BGL & \BGL
& \BGL & \BGL & \BGL & \BGL \\

Base & & 4.82 & 38.60\% & ---
     & & 6.95 & 14.00\% & ---
     & & 12.96 & 67.50\% & --- \\
KGW & & 5.59 & 27.30\% & 92.70\%
    & & 6.00 & 14.00\% & 93.30\%
    & & 13.00 & 57.50\% & 87.12\% \\
AM-F & & 9.86 & 29.30\% & 100.00\%
     & & 8.50 & 12.00\% & 100.00\%
     & & 15.69 & 32.50\% & 100.00\% \\
AM-F$^*$ & & --- & --- & 40.30\%
         & & --- & --- & 24.10\%
         & & --- & --- & 40.00\% \\
\DP DASMark & \DP & \DP 5.29 & \DP 32.00\% & \DP 94.40\%
            & \DP & \DP 6.33 & \DP 15.00\% & \DP 94.40\%
            & \DP & \DP 12.96 & \DP 67.50\% & \DP 98.50\% \\

\midrule
\BGL{\large\bfseries\itshape A-MEM}
& \BGL & \BGL & \BGL & \BGL
& \BGL & \BGL & \BGL & \BGL
& \BGL & \BGL & \BGL & \BGL \\

Base & & 4.77 & 31.70\% & ---
     & & 5.95 & 13.00\% & ---
     & & 13.07 & 70.00\% & --- \\
KGW & & 5.42 & 25.30\% & 90.00\%
    & & 5.55 & 11.00\% & 93.90\%
    & & 12.88 & 62.50\% & 89.60\% \\
AM-F & & 8.06 & 23.30\% & 100.00\%
     & & 4.60 & 15.00\% & 100.00\%
     & & 14.80 & 37.50\% & 100.00\% \\
AM-F$^*$ & & --- & --- & 25.10\%
         & & --- & --- & 20.20\%
         & & --- & --- & 54.50\% \\
\DP DASMark & \DP & \DP 5.08 & \DP 34.00\% & \DP 94.10\%
            & \DP & \DP 5.87 & \DP 15.00\% & \DP 94.30\%
            & \DP & \DP 13.07 & \DP 75.00\% & \DP 94.40\% \\

\midrule
\BGL{\large\bfseries\itshape Memento}
& \BGL & \BGL & \BGL & \BGL
& \BGL & \BGL & \BGL & \BGL
& \BGL & \BGL & \BGL & \BGL \\

Base & & 5.50 & 36.90\% & ---
     & & 6.76 & 15.00\% & ---
     & & 13.07 & 67.50\% & --- \\
KGW & & 5.57 & 23.30\% & 91.70\%
    & & 7.06 & 16.00\% & 96.20\%
    & & 12.17 & 57.50\% & 91.20\% \\
AM-F & & 9.00 & 25.30\% & 100.00\%
     & & 6.20 & 10.00\% & 100.00\%
     & & 16.14 & 35.00\% & 100.00\% \\
AM-F$^*$ & & --- & --- & 40.30\%
         & & --- & --- & 17.50\%
         & & --- & --- & 59.60\% \\
\DP DASMark & \DP & \DP 5.74 & \DP 36.00\% & \DP 92.40\%
            & \DP & \DP 7.12 & \DP 16.00\% & \DP 94.30\%
            & \DP & \DP 12.93 & \DP 67.50\% & \DP 98.40\% \\

\midrule
\BGG{\large\bfseries\itshape Evolving Harness}
& \BGG & \BGG & \BGG & \BGG
& \BGG & \BGG & \BGG & \BGG
& \BGG & \BGG & \BGG & \BGG \\

Base & & 4.82 & 28.90\% & ---
     & & 6.22 & 12.00\% & ---
     & & 12.93 & 57.50\% & --- \\
KGW & & 2.58 & 16.70\% & 87.20\%
    & & 5.42 & 12.00\% & 96.30\%
    & & 12.78 & 57.50\% & 80.65\% \\
AM-F & & 10.27 & 24.70\% & 100.00\%
     & & 6.20 & 15.00\% & 100.00\%
     & & 16.75 & 50.00\% & 100.00\% \\
AM-F$^*$ & & --- & --- & 36.10\%
         & & --- & --- & 11.60\%
         & & --- & --- & 70.50\% \\
\DP DASMark & \DP & \DP 4.93 & \DP 34.00\% & \DP 91.60\%
            & \DP & \DP 4.09 & \DP 11.00\% & \DP 93.80\%
            & \DP & \DP 13.23 & \DP 75.00\% & \DP 96.80\% \\

\bottomrule
\end{tabular}%
}

\caption{Watermark recovery and task utility across three domains and four runtime backends. Steps denotes average successful-task trajectory length, SR denotes task success rate, and Bit Acc. denotes bit accuracy. AM-F$^*$ uses re-elicited probabilities; ``---'' indicates an inapplicable metric.}
\label{tab:main-results}
\end{table*}
This section evaluates DASMark through three experiments.
Identity recovery experiments in dynamic environments
(Section~\ref{sec:experimental-setup}) assess the reliability
of APE and its impact on task utility.
Robustness experiments (Section~\ref{sec:robustness})
examine watermark recovery under trajectory perturbations
and the effectiveness of receiver enhancement.
Finally, ablation experiments (Section~\ref{sec:ablation})
evaluate the effectiveness of SRR in mitigating
context-induced decoding score shifts.

\subsection{Experimental Setup}
\label{sec:experimental-setup}
\begin{table*}[t]
\centering
\small
\setlength{\tabcolsep}{5pt}
\renewcommand{\arraystretch}{1.18}
\setlength{\aboverulesep}{0pt}
\setlength{\belowrulesep}{0pt}

\resizebox{\textwidth}{!}{%
\begin{tabular}{l*{9}{c}}
\toprule

\multirow{2}{*}{\textbf{Method}}
& \multicolumn{3}{c}{\textbf{Telecom}}
& \multicolumn{3}{c}{\textbf{CRMArena-Pro}}
& \multicolumn{3}{c}{\textbf{Retail}} \\

\cmidrule(lr){2-4}
\cmidrule(lr){5-7}
\cmidrule(lr){8-10}

& \textbf{Rew.} & \textbf{Del.} & \textbf{Syn.}
& \textbf{Rew.} & \textbf{Del.} & \textbf{Syn.}
& \textbf{Rew.} & \textbf{Del.} & \textbf{Syn.} \\

\midrule

AM-F
& 21.50\% & 27.87\% & 35.33\%
& 74.30\% & 2.71\%  & 11.03\%
& 42.25\% & 28.00\% & 29.17\% \\

KGW
& \underline{58.90\%} & 69.51\% & 79.39\%
& 79.28\% & 78.55\% & 84.02\%
& 66.20\% & 68.23\% & 76.99\% \\

\rowcolor{dasGold}
DAS
& 58.05\%
& \underline{77.62\%}
& \textbf{82.82\%}
& \underline{88.20\%}
& \underline{80.22\%}
& \underline{85.56\%}
& \underline{74.62\%}
& \underline{79.32\%}
& \underline{80.78\%} \\

\rowcolor{dasGold}
DAS-R
& \textbf{61.05\%}
& \textbf{79.18\%}
& \underline{81.79\%}
& \textbf{89.05\%}
& \textbf{82.93\%}
& \textbf{87.23\%}
& \textbf{77.40\%}
& \textbf{87.12\%}
& \textbf{87.48\%} \\

\bottomrule
\end{tabular}%
}

\caption{
Bit accuracy under trajectory perturbations, averaged over four runtime backends.
Rew., Del., and Syn. denote semantic rewriting, token deletion, and synonym substitution, respectively.
Del. and Syn. are averaged over attack rates of 10--60\%.
DAS-R uses a robustness-enhanced receiver.
}
\label{tab:robustness}
\end{table*}
\noindent\textbf{Environments and Tasks.}
The experiments include 150, 40, and 100 test tasks
from $\tau^3$-bench Telecom, $\tau^3$-bench Retail \citep{yao2024taubench,barres2025tau2},
and CRMArena-Pro
\citep{huang-etal-2026-crmarena-pro}, respectively. Each domain is evaluated under four runtime backends: No-Mem operates without memory, A-MEM \citep{xu2025amem} continuously updates memory, Memento \citep{zhou2025memento} evolves retrievable historical cases, and Evolving Harness \citep{cheng2026safecoevo} derives and updates prompting rules and skills from completed trajectories. All methods use identical tasks and execution orders, with watermark components frozen during evaluation.

\noindent\textbf{Baselines.}
DASMark is compared against the unwatermarked agent (Base) and two representative watermarking methods: KGW \citep{kirchenbauer2023watermark}, which embeds watermarks through token-level biasing, and AgentMark-F \citep{huang2026agentmark}, which employs behavior-based sampling. The comparison evaluates watermark recovery performance and task utility.

AgentMark-F relies on candidate action probabilities recorded during embedding for decoding. To assess its performance without access to these probabilities, an additional variant, AgentMark-F$^{*}$, is introduced. This variant preserves the original trajectories, actions, elicitation prompts, and secret keys, but decodes using candidate action probabilities re-estimated during verification. For a fair comparison, the payload sizes of KGW and AgentMark-F do not exceed the actual payload embedded by DASMark for each task.

\noindent\textbf{Evaluation Metrics.}
Bit accuracy (Bit Acc.) measures the proportion of correctly recovered bits among all embedded bits. Task success rate (SR) measures task utility, while Steps denotes the average number of decision steps across successful tasks.

\subsection{Identity Recovery in Dynamic Environments}
\label{sec:identity_recovery}

Memory updates, accumulated experience, and changes in runtime configurations in dynamic environments may shift the distribution of agent decision expressions, affecting the stability of identity watermark recovery. Table~\ref{tab:main-results} compares the watermark recovery performance and task utility of DASMark across three task domains and four runtime backends. Across all 12 experimental settings, DASMark achieves an average bit recovery accuracy of 94.8\%, outperforming KGW by 4.0 percentage points. Compared with the No-Mem backend, its average accuracy decreases by only 1.3 percentage points under dynamic runtime backends.

Dependence on embedding-time auxiliary information represents another important distinction between DASMark and existing methods. AgentMark-F achieves 100\% bit recovery accuracy when candidate action probabilities from the embedding stage are available, but this requires access to the corresponding historical probability information during verification. When the original probabilities are unavailable and must be re-estimated, AgentMark-F* achieves only 11.6\%--70.5\% accuracy, revealing its sensitivity to the availability of probability information. In contrast, DASMark recovers identity bits directly from the final decision-expression trajectories using a learned receiver and relative score comparison, without storing, sharing, or reconstructing embedding-time action probability distributions. This eliminates the need for additional historical probability records during identity verification, making DASMark more suitable for agent systems with continuously evolving runtime states.

Regarding task utility, as shown in Table~\ref{tab:main-results}, DASMark maintains task success rates and execution trajectory lengths generally comparable to those of the unwatermarked Base across different runtime backends. DASMark embeds identity information only in decision expressions without modifying predetermined tool actions or invocation parameters, thereby reducing interference with the agent's original task execution.

\subsection{Robustness Analysis}
\label{sec:robustness}

This section evaluates the watermark recovery robustness of DASMark
and the effectiveness of receiver enhancement training through three
types of trajectory perturbations, including semantic rewriting,
token deletion, and synonym substitution. In this experiment,
AM-F is configured with access to the behavior distribution table
generated during watermark embedding.

As shown in Table~\ref{tab:robustness}, DASMark outperforms KGW in
eight of the nine experimental settings, achieving an average
improvement of 5.1 percentage points in bit decoding accuracy.
DASMark-R further improves the average accuracy by 2.9 percentage
points and consistently outperforms both baseline methods across all
settings. Detailed results are provided in Appendix~\ref{app:full-robustness}.

These results demonstrate that DASMark maintains reliable watermark
recovery under various trajectory perturbations. The distinguishable
decision-expression channels learned by APE reduce the dependence of
watermark signals on specific lexical patterns, while SRR mitigates
the influence of correlated score shifts through relative decoding.
Although AM-F has access to the probability table generated during
embedding, it still fails to achieve reliable recovery under common
attacks. Furthermore, the improvement of DASMark-R shows that recovery
accuracy can be further enhanced by strengthening receiver robustness
without modifying the embedding mechanism, providing support for
reliable identity attribution in dynamic environments.
\begin{figure}[t]
    \centering
    \includegraphics[width=\columnwidth]{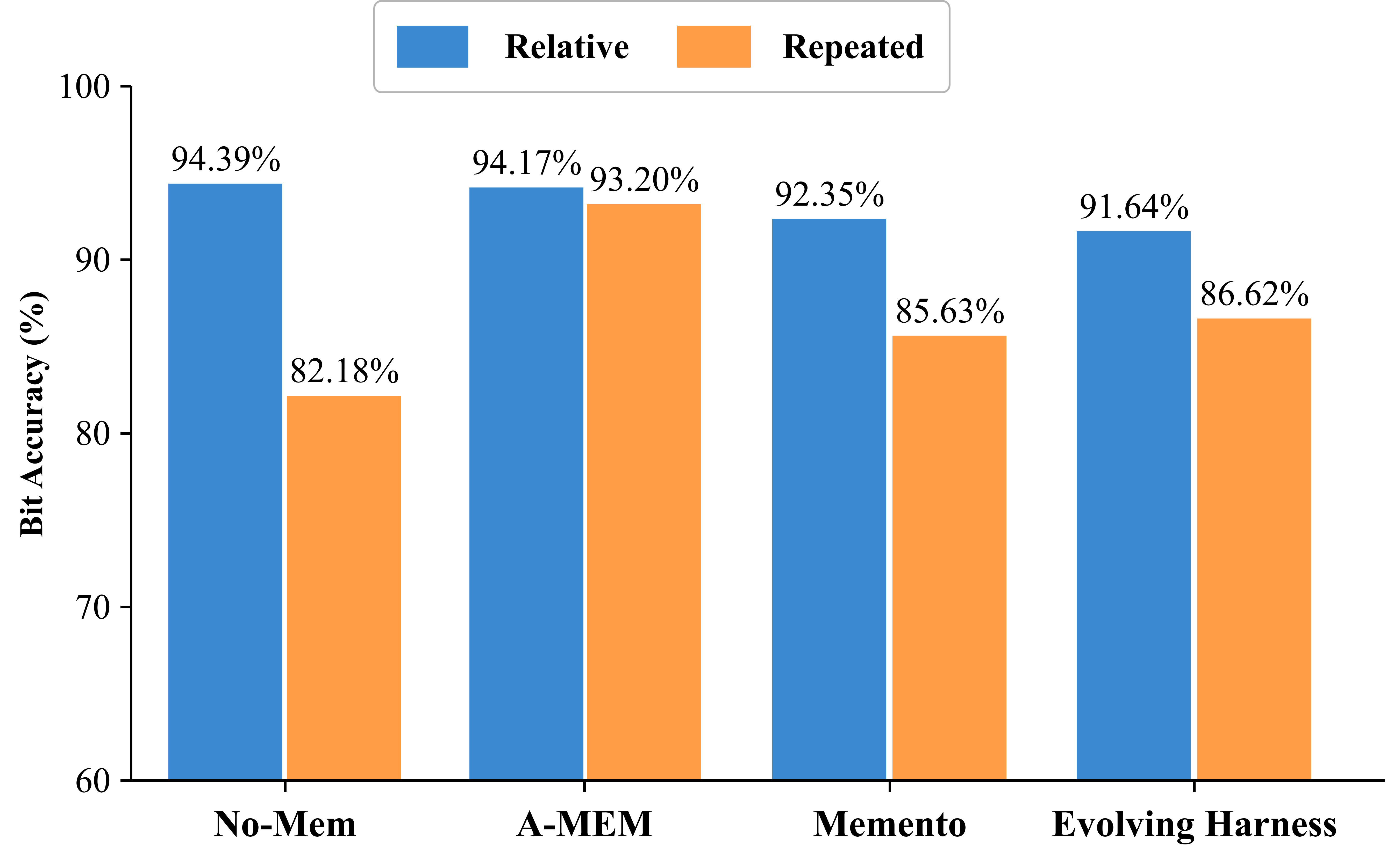}
    \caption{Bit accuracy of relative and repeated encoding and decoding strategies across four runtime backends on Telecom. Both strategies use two covers per bit.}
    \label{fig:srr-ablation}
\end{figure}
\subsection{Ablation Study}
\label{sec:ablation}

This section evaluates the effectiveness of Shift-Robust
Recovery (SRR) by comparing relative encoding and decoding
with repeated encoding and fixed-threshold decoding on
150 Telecom test tasks. Both strategies use two carriers
per bit, ensuring identical carrier overhead.

As shown in Figure~\ref{fig:srr-ablation}, the relative
encoding and decoding strategy consistently outperforms
repeated encoding and decoding across all four runtime
backends, achieving an average improvement of 6.23
percentage points in bit accuracy. Since both strategies
use the same number of carriers, the performance gain
mainly comes from the decoding strategy rather than
additional embedding redundancy.

These results demonstrate that DASMark maintains reliable watermark
recovery under various trajectory perturbations. The distinguishable
decision-expression channels learned by APE reduce the dependence of
watermarks on specific lexical patterns, while SRR mitigates the impact
of correlated score shifts through relative decoding. Although AM-F
has access to embedding-time probability information, it still fails to
achieve reliable recovery under common attacks. Furthermore, the
improvement of DASMark-R shows that recovery accuracy can be further
enhanced by improving receiver robustness without modifying the
watermark embedding mechanism, providing support for reliable identity
attribution in dynamic environments.

\section{Conclusion}
\label{sec:conclusion}

This paper presents DASMark, an identity watermarking
framework for agent systems in dynamic environments.
By combining action-preserving embedding with
shift-robust recovery, DASMark supports reliable
identity recovery as agent contexts evolve.
Across multiple task domains and dynamic backends,
DASMark improves bit-decoding accuracy under attacks
by an average of 5.1 percentage points over
state-of-the-art baselines, while maintaining task
performance comparable to unwatermarked agents.

\section*{Limitations}

This study primarily evaluates DASMark on agent task datasets
centered around tool invocation, with relatively limited
diversity in task types.
Future work will extend the evaluation to a broader range
of datasets and agent application scenarios to further
examine the generalizability of DASMark across diverse
environments.
Meanwhile, the current study mainly focuses on common
trajectory perturbations, including semantic rewriting,
token deletion, and synonym substitution, without fully
considering more sophisticated adversarial attacks.
Future work will investigate adaptive watermark removal,
identity forgery, and targeted trajectory manipulation
to further assess the security and robustness of DASMark
under complex adversarial conditions.

\bibliography{custom}
\newpage
\appendix

\section{Method and Implementation Details}
\label{app:implementation}

\subsection{Watermark Signal Construction}
\label{app:watermark-signal}

This section provides implementation details of state collection,
candidate generation, semantic filtering, and shared expression
direction estimation described in Section~3.2.

\paragraph{State Collection.}
The states preceding the first, middle, and last
valid tool decisions are extracted from real task trajectories
generated by the base agent. Complete interaction histories,
runtime contexts, and tool invocation information are preserved,
while duplicate or abnormal records are removed.
Each task and its derived states are assigned to the same
data partition. Test tasks are excluded from watermark
signal construction.

\paragraph{Candidate Generation and Filtering.}
For each state, the tool name and invocation arguments
are fixed, and Qwen3.5-9B \citep{qwen3.5} is used to
generate semantically equivalent decision expressions.
Eight candidates per state are sampled for Telecom and
CRMArena-Pro, and 16 for Retail, using a temperature
of 0.7 and deterministic random seeds derived from
state identifiers.

Generated expressions undergo normalization,
deduplication, and consistency checks.
Invalid outputs and expressions inconsistent with
the predetermined tool action, invocation arguments,
or runtime context are excluded.
The entire procedure uses only information visible
in the current state.

\paragraph{Semantic Encoding and Clustering.}
The frozen SentenceTransformer encoder
\texttt{all-MiniLM-L6-v2}
\citep{reimers-gurevych-2019-sentence}
is used to obtain semantic representations of candidate expressions.
The encoder accepts up to 256 tokens and produces
384-dimensional embeddings through attention-mask
mean pooling followed by $\ell_2$ normalization.

Candidates are clustered only within the same state.
Each retained cluster must have a minimum pairwise
cosine similarity of 0.85, and the ratio between
the longest and shortest expressions must not
exceed 1.5 in word count.
States with fewer than four valid candidates
are excluded.

\paragraph{Shared Expression Direction Estimation.}
The mean candidate embedding is first computed
for each state, and singular value decomposition
is then applied to the centered matrix of state means
to identify state-dependent directions.

For Telecom and CRMArena-Pro, the top 16 principal
directions are removed.
For Retail, the removed subspace is determined
by numerical rank using a singular-value threshold
of $10^{-8}$, resulting in 213 removed directions.

For candidate $j$ in state $i$, the state mean
is subtracted, and the centered embedding is
projected onto the orthogonal complement of
the state-dependent subspace:
\begin{equation}
r_{ij}
=
(I-V_rV_r^\top)(z_{ij}-\mu_i),
\end{equation}
where $z_{ij}$ denotes the normalized candidate
embedding, $\mu_i$ is the corresponding state mean,
and $V_r$ contains the removed directions.

To prevent states with more candidates from
dominating direction estimation, residual
covariances are first averaged within each
state and then equally aggregated across states:
\begin{equation}
C_{\mathrm{res}}
=
\frac{1}{N}
\sum_{i=1}^{N}
\frac{1}{n_i}
\sum_{j=1}^{n_i}
r_{ij}r_{ij}^{\top},
\end{equation}
where $N$ is the number of retained states
and $n_i$ is the number of valid candidates
in state $i$.

The shared expression direction is defined
as the unit eigenvector associated with the
largest eigenvalue of the residual covariance:
\begin{equation}
v^\ast
=
\underset{\|v\|_2=1}{\arg\max}
\;v^\top C_{\mathrm{res}}v.
\end{equation}

To ensure consistent channel assignments,
the coordinate of $v^\ast$ with the largest
absolute magnitude is constrained to be positive.

Each candidate is then assigned a projection
score $\alpha_{ij}=r_{ij}^{\top}v^\ast$.
Within each state, the two candidates with
the lowest scores and the two with the highest
scores are selected as supervision examples
for the two decision-expression channels,
respectively.

Only states with a positive separation margin
between the two candidate groups are retained.
Selected expressions must also pass the final
action, parameter, and factual consistency checks.

The semantic encoder, state-dependent subspace,
and filtering criteria are fixed before sender
training and are not adjusted using validation
or test decoding results.

\subsection{Sender and Receiver Training}
\label{app:training-details}

This section provides the training details of
Action-Preserving Embedding (APE) and
Shift-Robust Recovery (SRR), introduced in
Sections~3.3 and~3.4, respectively.

\paragraph{Sender Architecture and Configuration.}
DASMark builds two independent LoRA branches
$G_0$ and $G_1$ \citep{hu2022lora} on a frozen
Qwen3.5-9B backbone \citep{qwen3.5}.
Only the LoRA parameters in the \texttt{q\_proj}
and \texttt{v\_proj} modules are optimized.
For a base weight matrix $W$, branch
$c\in\{0,1\}$ computes
\begin{equation}
h_c = Wh + \frac{\alpha}{r}B_cA_ch.
\end{equation}

The LoRA rank is set to $r=16$, the scaling factor
to $\alpha=32$, and dropout to zero.
$A_c$ is initialized using Kaiming uniform
initialization \citep{he2015delving}, while $B_c$ is initialized to zero.

Both branches receive identical interaction histories,
runtime contexts, and predetermined tool actions,
but are supervised with different target decision
expressions.
Target expressions are limited to 192 tokens,
with a maximum context length of 8,192 tokens.
Invalid or overlength samples are excluded
without truncation.
Explicit thinking output is disabled, and
random seeds are fixed.

\paragraph{Sender Training.}
The sender jointly optimizes a text-fitting loss
and a channel separation objective:
\begin{equation}
\mathcal{L}_{\mathrm{send}}
=
\mathcal{L}_{\mathrm{fit}}
+
\lambda\tau\operatorname{softplus}
\left(
\frac{\gamma-g_b(y)}{\tau}
\right),
\end{equation}
where
$g_b(y)=\ell_b(y\mid s,a)-\ell_{1-b}(y\mid s,a)$
is the difference in average token log-likelihood
between the two branches.
$\mathcal{L}_{\mathrm{fit}}$ denotes the average
token negative log-likelihood of the target expression.
The parameters $\gamma=0.15$, $\lambda=2$, and $\tau=0.05$
are set to balance expression fidelity and channel
discriminability.
No receiver feedback or reinforcement learning
is used during sender training.

The backbone is loaded in bfloat16, while LoRA
parameters are optimized in float32.
AdamW is used
with a learning rate of $2\times10^{-5}$,
zero weight decay, a global batch size of 8,
and gradient clipping at 1.0.
Training runs for two epochs with gradient
checkpointing enabled.
Both LoRA branches are frozen after training.

\paragraph{Receiver Data and Architecture.}
The trained sender branches generate decision
expressions from interaction states in the training
and independent validation partitions.
Generation uses greedy decoding with a maximum
length of 192 tokens.
Only generated texts are used; candidate expressions
from watermark direction estimation are excluded.
Data are partitioned by task to prevent leakage,
and each valid state contributes one sample
per channel to maintain class balance.

Telecom and CRMArena-Pro use batches of 16 states
(32 texts), while Retail uses balanced batches
of 32 texts.

The receiver takes only the decision expression
as input, without task identifiers, tool metadata,
or runtime backend information.
Each text is converted into UTF-8 bytes and
processed by a 32-dimensional embedding layer
and a bidirectional GRU \citep{cho-etal-2014-learning}
with 64 hidden units per direction.

Masked mean, max, and attention pooling
are concatenated into a 384-dimensional
representation, followed by a 64-dimensional
linear layer, GELU activation, and a scalar
output layer producing the raw score $f_\phi(y)$.
Inputs exceeding 4,096 UTF-8 bytes are excluded
without truncation.

\paragraph{Receiver Training and Checkpoint Selection.}
The receiver is trained using binary cross-entropy,
with the true channel index as the supervision label.
Optimization uses AdamW with a learning rate
of $10^{-3}$, zero weight decay, and a gradient
clipping threshold of 1.0 for up to 60 epochs.
Validation is performed after each epoch on a
task-disjoint validation set.

Because online recovery relies on relative scores,
validation evaluates both single-cover
classification accuracy and same-state pairwise
ranking accuracy.
For a validation set containing $M$ states,
the latter is defined as
\begin{equation}
\begin{aligned}
\operatorname{Acc}_{\mathrm{pair}}
&=
\frac{1}{M}\sum_{i=1}^{M}
\mathbb{I}\bigl[
f_\phi(y_i^1)
\\
&\qquad >
f_\phi(y_i^0)
\bigr],
\end{aligned}
\end{equation}
where $y_i^0$ and $y_i^1$ are the expressions
generated by the two sender branches for
the same state.

Checkpoint selection follows predefined,
domain-specific criteria.
For CRMArena-Pro, pairwise ranking accuracy
is the primary criterion, followed by
single-cover classification accuracy.
For Telecom, balanced classification accuracy
is prioritized, followed by overall accuracy.
For Retail, checkpoints are selected based on
accuracy under dynamic contexts and overall
accuracy on predefined validation tasks.
When validation results are identical across
the applicable criteria, the earlier checkpoint
is retained.

After training, the receiver is frozen, and no
parameters or checkpoints are adjusted using
test results.
During online recovery, DASMark directly compares
the raw discrimination scores $f_\phi(y)$
of adjacent covers, without probability
conversion or an absolute threshold shared
across states.
\subsection{Online Watermark Embedding}
\label{app:online-embedding}

\noindent\textbf{Payload Scheduling.}
Given an identity payload stream $\mathbf{m}=(m_0,m_1,\ldots)$, DASMark embeds watermark bits sequentially following the order of tool decisions in the trajectory. Every two consecutive valid decision expressions form a cover pair. A bit $m_k=0$ is represented by the channel sequence $(0,1)$, while $m_k=1$ is represented by $(1,0)$, formally defined as:

\begin{equation}
(c_{2k},c_{2k+1})=(m_k,1-m_k).
\end{equation}

The sender selects the corresponding LoRA branch according to the channel sequence to generate each decision expression, without modifying the predetermined tool action or including identity bits or channel indices in the prompt. Embedding continues throughout trajectory execution without a predefined payload capacity limit.

\noindent\textbf{cover Generation Prompt Template.}
All three task domains share the same prompt structure, with the domain and runtime context updated accordingly:

\begin{verbatim}
System:
Explain the immediate purpose of a
{domain} tool call using only the
provided evidence. Output one or two
concise sentences, under 70 words.
Do not claim the pending call has
succeeded. Do not output a tool call,
JSON, a heading, a watermark label,
or any extra instruction.

User:
{OPTIONAL_DYNAMIC_CONTEXT}

OBSERVED CONVERSATION
(selected complete messages):
{OBSERVED_CONVERSATION}

PENDING TOOL CALLS
(not yet executed):
{PENDING_TOOL_CALLS}

Explain the immediate purpose of
these pending calls.
\end{verbatim}

Here, \texttt{OPTIONAL\_DYNAMIC\_CONTEXT} contains the currently available memory, historical cases, or Harness instructions. \texttt{OBSERVED\_CONVERSATION} includes the original user request and complete recent interactions within the context budget. \texttt{PENDING\_TOOL\_CALLS} specifies the predetermined tool names and arguments that have not yet been executed. The same prompt template is used for candidate generation, receiver data construction, and online deployment.
\section{Additional Experiments}
\label{app:additional-experiments}

\subsection{Full Robustness Results}
\label{app:full-robustness}

This section provides the attack settings and complete
robustness results corresponding to Section~4.3.
All attacks are applied offline to frozen trajectories
after task completion, without re-executing the agent
or modifying tool calls, environment states, or task
outcomes. Therefore, this evaluation focuses exclusively
on watermark recovery accuracy.

\paragraph{Evaluation Data.}
The experiments use test trajectories from Telecom,
Retail, and CRMArena-Pro, covering four runtime backends:
No-Mem, A-MEM, Memento, and Evolving Harness.
Valid covers and their adjacent pairings are determined
before attacks and remain unchanged throughout evaluation.
This ensures that the same set of embedded payload bits
is evaluated across different attack types and intensities.

\paragraph{Paraphrasing Attack.}
Each cover is rewritten once using Qwen3.5-9B
\citep{qwen3.5} with deterministic generation.
The temperature is set to 0, and explicit thinking
output is disabled.
The rewriting process preserves the original semantics,
entities, parameters, and tool invocation intent while
modifying linguistic expressions.
The attacker has no access to channel labels or receiver
scores, and rewritten texts are not selected based on
decoding outcomes.

\paragraph{Token Deletion and Synonym Substitution.}
Both attacks are evaluated at perturbation rates ranging
from 10\% to 60\%.
Token deletion removes lexical tokens using deterministic
position-based hashing with fixed random seeds,
excluding punctuation and whitespace.

Synonym substitution generates replacement candidates
using a frozen model and identifies eligible positions
through word-sequence alignment.
The substitution rate is calculated over positions
with valid candidates rather than all tokens.
Both attacks employ deterministic nested sampling,
ensuring that positions perturbed at lower attack rates
remain included at higher rates.

\paragraph{Post-Attack Decoding.}
After each attack, the frozen receiver recomputes
cover scores and performs differential recovery
using the original adjacent cover pairings.
Invalid or unscorable covers are counted as
bit recovery failures rather than excluded.
Bit accuracy is computed over all originally
embedded payload bits to ensure comparability
across attack settings.

\paragraph{Robustness-Enhanced Receiver.}
DASMark-R shares the same sender and watermark
embedding procedure as standard DASMark,
but enhances receiver training with perturbed samples.
Training data include paraphrased and locally deleted
expressions, while preserving their original channel labels.
Augmented training excludes test data, and checkpoints
are selected exclusively using an independent
validation partition.

\paragraph{Complete Results.}
The complete robustness results are reported in three
tables, corresponding to Telecom, Retail, and
CRMArena-Pro, respectively.

For Telecom, Table~\ref{tab:telecom-robustness}
presents the results under
\hyperref[tab:telecom-semantic-rewriting]{(a) paraphrasing},
\hyperref[tab:telecom-token-delete]{(b) token deletion},
and
\hyperref[tab:telecom-synonym-replacement]{(c) synonym substitution}.

For Retail, Table~\ref{tab:retail-robustness}
reports the corresponding results under
\hyperref[tab:retail-semantic-rewriting]{(a) paraphrasing},
\hyperref[tab:retail-token-delete]{(b) token deletion},
and
\hyperref[tab:retail-synonym-replacement]{(c) synonym substitution}.

For CRMArena-Pro,
Table~\ref{tab:CRMArena-Pro-robustness}
presents the results under
\hyperref[tab:CRMArena-Pro-semantic-rewriting]{(a) paraphrasing},
\hyperref[tab:CRMArena-Pro-token-delete]{(b) token deletion},
and
\hyperref[tab:CRMArena-Pro-synonym-replacement]{(c) synonym substitution}.

All three tables cover the four runtime backends,
allowing comparison across different attack types,
perturbation intensities, and dynamic runtime contexts.

Overall, the three attacks affect watermark covers
by modifying linguistic structures, reducing available
textual evidence, or perturbing local lexical choices.
Dynamic runtime backends further introduce variations
in the text distributions encountered by the receiver.

DASMark-R generally achieves more stable recovery
under high-intensity perturbations, indicating that
enhancing receiver robustness to perturbed expressions
can mitigate attack-induced degradation without
modifying the sender or embedding procedure.

\subsection{Transferability and Model Replacement}
\label{app:transferability}

This section evaluates the applicability of DASMark
under cross-domain transfer and base model replacement.
These results are not included in the average in-domain
results reported in the main text.
\begin{table}[t]
\centering
\small
\setlength{\tabcolsep}{10pt}
\setlength{\extrarowheight}{1.5pt}
\renewcommand{\arraystretch}{1.1}
\setlength{\aboverulesep}{0pt}
\setlength{\belowrulesep}{0pt}

\begin{tabular}{llcc}
\toprule
\rowcolor{hdrGray}
\textbf{Dataset}
& \textbf{Metric}
& \textbf{DASMark}
& \textbf{Base} \\
\midrule

\multirow{3}{*}{Retail}
& Steps $\downarrow$
& 13.18
& 13.15 \\

& Bit Acc $\uparrow$
& \textbf{98.5\%}
& -- \\

& SR $\uparrow$
& 67.5\%
& 67.5\% \\

\midrule

\multirow{3}{*}{ToolBench}
& Steps $\downarrow$
& 3.13
& 3.59 \\

& Bit Acc $\uparrow$
& \textbf{94.8\%}
& -- \\

& SR $\uparrow$
& 80.0\%
& 79.0\% \\

\bottomrule
\end{tabular}

\caption{OOD transfer evaluation from Retail to ToolBench. DASMark is trained
on Retail and directly transferred to 100 ToolBench tasks without additional
training. Steps denotes the average trajectory length, SR denotes task success
rate, and Bit Acc denotes watermark decoding accuracy.}
\label{tab:ood-transfer}
\end{table}

\paragraph{Cross-Domain Transfer.}
Retail is used as the source domain, and 100
retail-related tasks are selected from ToolBench
\citep{qin2024toolllm} for out-of-distribution (OOD)
evaluation.
Both the sender and receiver are trained on Retail
and remain frozen during transfer, without additional
training or parameter adaptation using target-domain data.

Table~\ref{tab:ood-transfer} shows that DASMark
maintains high watermark recovery accuracy on
related-domain tasks, while preserving task utility
and trajectory length comparable to the unwatermarked
agent. This suggests that the learned decision-expression
channels exhibit a degree of cross-task transferability,
although the results do not establish generalization
to arbitrary unseen domains.
\begin{table}[t]
\centering
\small
\setlength{\tabcolsep}{3pt}
\renewcommand{\arraystretch}{1.12}
\setlength{\aboverulesep}{0pt}
\setlength{\belowrulesep}{0pt}

\resizebox{\columnwidth}{!}{%
\begin{tabular}{lcccc}
\toprule

\textbf{Method}
& \textbf{No-Mem}
& \textbf{A-MEM}
& \textbf{Memento}
& \textbf{Harness} \\

\midrule

AM-F
& 17.20\%
& 10.40\%
& 13.60\%
& 5.80\% \\

KGW
& \textbf{59.40\%}
& 55.60\%
& 58.50\%
& 59.50\% \\

\rowcolor{dasGold}
DASMark
& 54.80\%
& \textbf{70.70\%}
& \textbf{83.80\%}
& \textbf{67.40\%} \\

\bottomrule
\end{tabular}%
}

\caption{
Bit accuracy of Qwen3.5-4B under two rounds of semantic rewriting
across four runtime backends using Qwen3.5-4B.
Bold values indicate the best performance in each backend.
}

\label{tab:semantic-rewriting}
\end{table}
\paragraph{Base Model Replacement.}
Qwen3.5-9B is replaced with Qwen3.5-4B, and
watermark signal construction, sender training,
and receiver training are repeated using the
same Telecom task partition.
The experiment includes 1,000 training tasks,
300 validation tasks, and 150 test tasks.
The watermark components are trained independently
for the two base models.

Table~\ref{tab:model-comparison} shows that DASMark
maintains high watermark recovery accuracy on the
smaller model, despite a substantial decrease in
task success rate. This indicates that watermark
recoverability and the underlying agent's task
execution capability are not necessarily aligned.
It also demonstrates that DASMark can be adapted
to different model scales through retraining.

\definecolor{dasGoldBright}{RGB}{255,230,150}
\definecolor{backendGray}{RGB}{236,238,241}

\begin{table*}[t]
\centering
\small
\setlength{\tabcolsep}{3pt}
\renewcommand{\arraystretch}{1.15}
\setlength{\aboverulesep}{0pt}
\setlength{\belowrulesep}{0pt}

\resizebox{\textwidth}{!}{%
\begin{tabular}{lccc@{\hskip 9pt}ccc@{\hskip 9pt}ccc}
\toprule

\multirow{2}{*}{\textbf{Method}}
& \multicolumn{3}{c}{\textbf{Qwen3.5-4B}}
& \multicolumn{3}{c}{\textbf{Qwen3.5-9B}}
& \multicolumn{3}{c}{\textbf{Performance Gap ($\Delta$: 4B $-$ 9B)}} \\

\cmidrule(lr){2-4}
\cmidrule(lr){5-7}
\cmidrule(lr){8-10}

& \textbf{Steps~$\downarrow$}
& \textbf{SR~$\uparrow$}
& \textbf{Bit Acc.~$\uparrow$}
& \textbf{Steps~$\downarrow$}
& \textbf{SR~$\uparrow$}
& \textbf{Bit Acc.~$\uparrow$}
& \textbf{Steps}
& \textbf{SR (pp)}
& \textbf{Bit Acc. (pp)} \\

\midrule


\rowcolor{backendGray}
\multicolumn{10}{l}{\textbf{\textit{No-Mem}}} \\

Base
& 4.43 & \textbf{4.67\%} & --
& \textbf{4.82} & \textbf{38.60\%} & --
& -0.39 & -33.93 & -- \\

KGW
& \textbf{4.10} & 6.67\% & 92.40\%
& 5.59 & 27.30\% & 92.70\%
& -1.49 & -20.63 & -0.30 \\

AM-F
& 7.33 & 2.00\% & \textbf{100.00\%}
& 9.86 & 29.30\% & \textbf{100.00\%}
& -2.53 & -27.30 & 0.00 \\

\rowcolor{dasGoldBright}
DASMark
& 5.29 & \textbf{8.67\%} & \textbf{100.00\%}
& 5.29 & 32.00\% & 94.40\%
& 0.00 & -23.33 & +5.60 \\

\midrule


\rowcolor{backendGray}
\multicolumn{10}{l}{\textbf{\textit{A-MEM}}} \\

Base
& 4.45 & \textbf{6.00\%} & --
& \textbf{4.77} & 31.70\% & --
& -0.32 & -25.70 & -- \\

KGW
& \textbf{3.83} & 4.00\% & 90.00\%
& 5.42 & 25.30\% & 90.00\%
& -1.59 & -21.30 & 0.00 \\

AM-F
& 6.20 & 3.33\% & \textbf{100.00\%}
& 8.06 & 23.30\% & \textbf{100.00\%}
& -1.86 & -19.97 & 0.00 \\

\rowcolor{dasGoldBright}
DASMark
& 5.08 & 4.67\% & \textbf{100.00\%}
& 5.08 & \textbf{34.00\%} & 94.10\%
& 0.00 & -29.33 & +5.90 \\

\midrule


\rowcolor{backendGray}
\multicolumn{10}{l}{\textbf{\textit{Memento}}} \\

Base
& \textbf{4.23} & 5.33\% & --
& \textbf{5.50} & \textbf{36.90\%} & --
& -1.27 & -31.57 & -- \\

KGW
& 4.71 & 4.67\% & 91.70\%
& 5.57 & 23.30\% & 91.70\%
& -0.86 & -18.63 & 0.00 \\

AM-F
& 8.00 & 0.67\% & \textbf{100.00\%}
& 9.00 & 25.30\% & \textbf{100.00\%}
& -1.00 & -24.63 & 0.00 \\

\rowcolor{dasGoldBright}
DASMark
& 5.74 & \textbf{6.00\%} & \textbf{100.00\%}
& 5.74 & 36.00\% & 92.40\%
& 0.00 & -30.00 & +7.60 \\

\midrule


\rowcolor{backendGray}
\multicolumn{10}{l}{\textbf{\textit{Evolving Harness}}} \\

Base
& \textbf{4.40} & 2.67\% & --
& 4.82 & 28.90\% & --
& -0.42 & -26.23 & -- \\

KGW
& 4.67 & \textbf{6.00\%} & 87.20\%
& \textbf{2.58} & 16.70\% & 87.20\%
& +2.09 & -10.70 & 0.00 \\

AM-F
& 8.50 & 1.33\% & \textbf{100.00\%}
& 10.27 & 24.70\% & \textbf{100.00\%}
& -1.77 & -23.37 & 0.00 \\

\rowcolor{dasGoldBright}
DASMark
& 4.93 & \textbf{6.00\%} & \textbf{100.00\%}
& 4.93 & \textbf{34.00\%} & 91.60\%
& 0.00 & -28.00 & +8.40 \\

\bottomrule
\end{tabular}%
}

\caption{
Comparison across Qwen3.5 model sizes under four runtime
backends. Steps denotes the average trajectory length
among successful tasks, SR denotes task success rate,
and Bit Acc. denotes watermark decoding accuracy.
The performance gap is computed as Qwen3.5-4B minus
Qwen3.5-9B, with accuracy differences reported in
percentage points (pp). Bold values indicate the best
performance for each metric within each backend
and model size, including ties.
}

\label{tab:model-comparison}
\end{table*}

\begin{table*}[t]
\centering
\small
\setlength{\tabcolsep}{5pt}
\renewcommand{\arraystretch}{1.15}
\setlength{\aboverulesep}{0pt}
\setlength{\belowrulesep}{0pt}

\resizebox{\textwidth}{!}{%
\begin{tabular}{llcccccc}
\toprule

\multirow{2}{*}{\textbf{Method}}
& \multirow{2}{*}{\textbf{Backend}}
& \multicolumn{6}{c}{\textbf{Token Deletion Rate}} \\
\cmidrule(lr){3-8}
& & \textbf{10\%} & \textbf{20\%}
  & \textbf{30\%} & \textbf{40\%}
  & \textbf{50\%} & \textbf{60\%} \\
\midrule

& No-Mem
& 68.80\% & 43.80\% & 27.60\%
& 16.10\% & 9.90\% & 5.20\% \\

& A-MEM
& 76.50\% & 51.30\% & 37.40\%
& 19.10\% & 8.70\% & 4.30\% \\

& Memento
& 75.80\% & 47.00\% & 30.30\%
& 19.70\% & 7.60\% & 6.10\% \\

\multirow{-4}{*}{\textbf{AgentMark-F}}
& Evolving Harness
& 71.20\% & 50.00\% & 36.50\%
& 23.10\% & 19.20\% & 1.90\% \\

\midrule

& No-Mem
& 91.60\% & 82.90\% & 75.50\%
& 66.80\% & 61.40\% & 58.70\% \\

& A-MEM
& 89.20\% & 83.90\% & 76.20\%
& 68.50\% & 63.30\% & 59.40\% \\

& Memento
& 84.80\% & 75.50\% & 73.30\%
& 68.20\% & 66.10\% & 62.10\% \\

\multirow{-4}{*}{\textbf{KGW}}
& Evolving Harness
& 91.00\% & 85.30\% & 77.90\%
& 71.60\% & 63.90\% & 57.90\% \\

\midrule

\rowcolor{dasGold}
& No-Mem
& \textbf{98.80\%} & \textbf{97.20\%}
& \textbf{93.30\%} & \textbf{89.70\%}
& \textbf{86.90\%} & \textbf{82.10\%} \\

\rowcolor{dasGold}
& A-MEM
& \textbf{97.90\%} & \textbf{95.40\%}
& \textbf{93.30\%} & \textbf{90.40\%}
& \textbf{85.40\%} & \textbf{91.20\%} \\

\rowcolor{dasGold}
& Memento
& \textbf{99.60\%} & \textbf{98.80\%}
& \textbf{95.80\%} & \textbf{90.40\%}
& \textbf{88.30\%} & \textbf{82.10\%} \\

\rowcolor{dasGold}
\multirow{-4}{*}{\textbf{DASMark}}
& Evolving Harness
& \textbf{99.60\%} & \textbf{97.50\%}
& \textbf{95.50\%} & \textbf{90.10\%}
& \textbf{85.50\%} & \textbf{78.10\%} \\

\bottomrule
\end{tabular}%
}

\caption{
Bit accuracy of Qwen3.5-4B under token deletion across four runtime
backends and deletion rates from 10\% to 60\%.
Bold values indicate the best performance for each
backend and deletion rate.
}
\label{tab:token-delete}
\end{table*}

\paragraph{Robustness After Model Replacement.}
Tables~\ref{tab:semantic-rewriting},
\ref{tab:token-delete} report the results
under semantic rewriting and token deletion
attacks, respectively.
Semantic rewriting is applied twice, while
deletion and substitution rates range from
10\% to 60\%.

The results show that DASMark retains strong
resistance to textual perturbations on the smaller
model, generally outperforming the baselines
under token deletion and synonym substitution.
However, structural changes introduced by semantic
rewriting still affect recovery under certain
runtime backends, and accuracy decreases under
stronger perturbations.

Overall, DASMark demonstrates adaptability to
related-domain tasks and different base model
scales, although its practical effectiveness
remains influenced by domain differences,
underlying model capabilities, and attack intensity.

\begin{table*}[p]
\centering
\footnotesize
\setlength{\tabcolsep}{4pt}
\renewcommand{\arraystretch}{1.0}
\setlength{\aboverulesep}{0pt}
\setlength{\belowrulesep}{0pt}

\textbf{(a) Paraphrasing (Semantic Rewriting)}\par\vspace{3pt}
\begin{tabular*}{\textwidth}{@{\extracolsep{\fill}}lcccc@{}}
\toprule

\textbf{Method}
& \textbf{No-Mem}
& \textbf{A-MEM}
& \textbf{Memento}
& \textbf{Evolving Harness} \\
\midrule

AM-F
& 29.80\%
& 28.50\%
& 11.80\%
& 15.90\% \\

KGW
& \textbf{54.90\%}
& 60.60\%
& 59.50\%
& 60.60\% \\

\rowcolor{dasGold}
DASMark
& 51.80\%
& 53.10\%
& \textbf{65.10\%}
& 62.20\% \\

\rowcolor{dasGold}
DASMark-R
& 54.40\%
& \textbf{61.20\%}
& 58.70\%
& \textbf{69.90\%} \\
\bottomrule
\end{tabular*}

\vspace{4pt}

\textbf{(b) Token Deletion}\par\vspace{3pt}
\resizebox{\textwidth}{!}{%
\begin{tabular}{llcccccc}
\toprule

\multirow{2}{*}{\textbf{Method}}
& \multirow{2}{*}{\textbf{Backend}}
& \multicolumn{6}{c}{\textbf{Token Deletion Rate}} \\

\cmidrule(lr){3-8}

& & \textbf{10\%} & \textbf{20\%}
  & \textbf{30\%} & \textbf{40\%}
  & \textbf{50\%} & \textbf{60\%} \\

\midrule
& No-Mem
& 62.90\% & 42.30\% & 25.00\%
& 13.70\% & 10.50\% & 6.00\% \\

& A-MEM
& 63.30\% & 43.40\% & 27.00\%
& 16.90\% & 9.40\% & 4.50\% \\

& Memento
& 74.10\% & 48.70\% & 32.30\%
& 20.50\% & 8.00\% & 4.60\% \\

\multirow{-4}{*}{\textbf{AgentMark-F}}
& Evolving Harness
& 70.80\% & 42.90\% & 18.00\%
& 12.90\% & 7.30\% & 3.90\% \\

\midrule


& No-Mem
& 85.40\% & 78.50\% & 73.60\%
& 69.80\% & 69.10\% & 59.00\% \\

& A-MEM
& 84.40\% & 78.90\% & 70.90\%
& 62.30\% & 57.10\% & 56.40\% \\

& Memento
& 84.40\% & 79.10\% & 70.20\%
& 62.30\% & 57.40\% & 58.00\% \\

\multirow{-4}{*}{\textbf{KGW}}
& Evolving Harness
& 82.80\% & 75.80\% & 69.00\%
& 64.00\% & 62.30\% & 57.60\% \\

\midrule

\rowcolor{dasGold}
& No-Mem
& 87.50\% & 85.50\% & \textbf{81.50\%}
& 74.30\% & 71.00\% & 64.00\% \\

\rowcolor{dasGold}
& A-MEM
& 91.20\% & \textbf{90.60\%} & 82.70\%
& 77.90\% & 74.60\% & 70.40\% \\

\rowcolor{dasGold}
& Memento
& 88.10\% & 83.50\% & 77.10\%
& 74.00\% & 68.80\% & 70.60\% \\

\rowcolor{dasGold}
\multirow{-4}{*}{\textbf{DASMark}}
& Evolving Harness
& \textbf{87.60\%} & \textbf{82.30\%} & 75.60\%
& 72.20\% & 66.60\% & 65.20\% \\


\rowcolor{dasGold}
& No-Mem
& \textbf{89.40\%} & \textbf{85.80\%} & 79.20\%
& \textbf{74.90\%} & \textbf{72.30\%} & \textbf{70.00\%} \\

\rowcolor{dasGold}
& A-MEM
& \textbf{91.90\%} & 89.60\% & \textbf{85.00\%}
& \textbf{80.50\%} & \textbf{75.20\%} & \textbf{70.70\%} \\

\rowcolor{dasGold}
& Memento
& \textbf{90.50\%} & \textbf{85.00\%} & \textbf{80.70\%}
& \textbf{77.70\%} & \textbf{76.50\%} & \textbf{71.90\%} \\

\rowcolor{dasGold}
\multirow{-4}{*}{\textbf{DASMark-R}}
& Evolving Harness
& 82.60\% & 81.60\% & \textbf{77.90\%}
& \textbf{72.60\%} & \textbf{71.20\%} & \textbf{67.60\%} \\

\bottomrule
\end{tabular}%
}

\vspace{4pt}

\textbf{(c) Synonym Substitution}\par\vspace{3pt}
\resizebox{\textwidth}{!}{%
\begin{tabular}{llcccccc}
\toprule

\multirow{2}{*}{\textbf{Method}}
& \multirow{2}{*}{\textbf{Backend}}
& \multicolumn{6}{c}{\textbf{Synonym Substitution Rate}} \\

\cmidrule(lr){3-8}

& & \textbf{10\%} & \textbf{20\%}
  & \textbf{30\%} & \textbf{40\%}
  & \textbf{50\%} & \textbf{60\%} \\

\midrule

& No-Mem
& 72.60\% & 54.80\% & 40.30\%
& 25.00\% & 15.30\% & 6.90\% \\

& A-MEM
& 77.90\% & 54.70\% & 33.00\%
& 19.50\% & 10.10\% & 4.90\% \\

& Memento
& 74.50\% & 55.90\% & 40.70\%
& 28.90\% & 11.40\% & 5.30\% \\

\multirow{-4}{*}{\textbf{AgentMark-F}}
& Evolving Harness
& 78.50\% & 53.20\% & 39.10\%
& 24.00\% & 14.60\% & 6.90\% \\

\midrule


& No-Mem
& 89.20\% & 84.40\% & 78.80\%
& 75.30\% & 75.00\% & 72.20\% \\

& A-MEM
& 86.50\% & 85.50\% & 82.40\%
& 78.50\% & 74.70\% & 70.90\% \\

& Memento
& 88.70\% & 85.00\% & \textbf{81.00\%}
& \textbf{78.80\%} & \textbf{77.30\%} & 72.10\% \\

\multirow{-4}{*}{\textbf{KGW}}
& Evolving Harness
& 85.90\% & 81.80\% & 79.10\%
& 74.10\% & 76.10\% & 72.10\% \\

\midrule
\rowcolor{dasGold}
& No-Mem
& \textbf{91.40\%} & \textbf{89.80\%}
& \textbf{86.50\%} & \textbf{83.20\%}
& 79.50\% & 75.20\% \\

\rowcolor{dasGold}
& A-MEM
& \textbf{92.50\%} & \textbf{90.90\%}
& 87.00\% & \textbf{86.30\%}
& \textbf{83.10\%} & \textbf{81.80\%} \\

\rowcolor{dasGold}
& Memento
& \textbf{89.90\%} & \textbf{85.60\%}
& \textbf{81.00\%} & 75.80\%
& 73.10\% & 70.90\% \\

\rowcolor{dasGold}
\multirow{-4}{*}{\textbf{DASMark}}
& Evolving Harness
& \textbf{86.30\%} & \textbf{83.30\%}
& \textbf{82.30\%} & \textbf{79.30\%}
& \textbf{79.30\%} & 73.60\% \\


\rowcolor{dasGold}
& No-Mem
& 88.40\% & 86.50\% & 85.10\%
& 80.50\% & \textbf{79.90\%}
& \textbf{76.20\%} \\

\rowcolor{dasGold}
& A-MEM
& 91.90\% & 90.20\% & \textbf{87.90\%}
& 83.40\% & 82.40\% & 81.40\% \\

\rowcolor{dasGold}
& Memento
& 85.90\% & 84.40\% & 79.80\%
& 78.30\% & 76.50\%
& \textbf{75.20\%} \\

\rowcolor{dasGold}
\multirow{-4}{*}{\textbf{DASMark-R}}
& Evolving Harness
& 83.90\% & 79.60\% & 77.30\%
& 76.30\% & 77.30\%
& \textbf{74.60\%} \\

\bottomrule
\end{tabular}%
}

\caption{Complete robustness results on Telecom under (a) paraphrasing, (b) token deletion (10--60\%), and (c) synonym substitution (10--60\%) across four runtime backends. Values denote bit-decoding accuracy. DASMark-R uses a robustness-enhanced receiver. Bold values mark the best accuracy for each backend and attack setting (including ties).}
\label{tab:telecom-robustness}
\label{tab:telecom-semantic-rewriting}
\label{tab:telecom-token-delete}
\label{tab:telecom-synonym-replacement}
\end{table*}

\begin{table*}[p]
\centering
\footnotesize
\setlength{\tabcolsep}{4pt}
\renewcommand{\arraystretch}{1.0}
\setlength{\aboverulesep}{0pt}
\setlength{\belowrulesep}{0pt}

\textbf{(a) Paraphrasing (Semantic Rewriting)}\par\vspace{3pt}
\begin{tabular*}{\textwidth}{@{\extracolsep{\fill}}lcccc@{}}
\toprule

\textbf{Method}
& \textbf{No-Mem}
& \textbf{A-MEM}
& \textbf{Memento}
& \textbf{Evolving Harness} \\

\midrule

AM-F
& 41.70\%
& 48.50\%
& 26.30\%
& 52.50\% \\

KGW
& 65.90\%
& 66.40\%
& \textbf{71.20\%}
& 61.30\% \\

\rowcolor{dasGold}
DASMark
& 81.10\%
& 73.60\%
& 68.80\%
& 75.00\% \\

\rowcolor{dasGold}
DASMark-R
& \textbf{81.80\%}
& \textbf{79.20\%}
& \textbf{71.20\%}
& \textbf{77.40\%} \\

\bottomrule
\end{tabular*}

\vspace{4pt}

\textbf{(b) Token Deletion}\par\vspace{3pt}
\resizebox{\textwidth}{!}{%
\begin{tabular}{llcccccc}
\toprule

\multirow{2}{*}{\textbf{Method}}
& \multirow{2}{*}{\textbf{Backend}}
& \multicolumn{6}{c}{\textbf{Token Deletion Rate}} \\

\cmidrule(lr){3-8}

& & \textbf{10\%} & \textbf{20\%}
  & \textbf{30\%} & \textbf{40\%}
  & \textbf{50\%} & \textbf{60\%} \\

\midrule

& No-Mem
& 68.30\% & 38.30\% & 23.30\%
& 15.00\% & 0.00\% & 0.00\% \\

& A-MEM
& 78.80\% & 57.60\% & 30.30\%
& 18.20\% & 3.00\% & 0.00\% \\

& Memento
& 77.20\% & 38.60\% & 21.10\%
& 17.50\% & 10.50\% & 7.00\% \\

\multirow{-4}{*}{\textbf{AgentMark-F}}
& Evolving Harness
& 72.10\% & 54.10\% & 27.90\%
& 13.10\% & 0.00\% & 0.00\% \\

\midrule


& No-Mem
& 81.80\% & 72.00\% & 69.70\%
& 65.20\% & 62.90\% & 61.40\% \\

& A-MEM
& 84.00\% & 78.40\% & 72.00\%
& 65.60\% & 64.80\% & 56.00\% \\

& Memento
& 86.40\% & 80.00\% & 74.40\%
& 66.40\% & 62.40\% & 58.40\% \\

\multirow{-4}{*}{\textbf{KGW}}
& Evolving Harness
& 79.80\% & 73.40\% & 62.90\%
& 52.40\% & 53.20\% & 54.00\% \\

\midrule

\rowcolor{dasGold}
& No-Mem
& 90.20\% & 82.60\% & 78.80\%
& 77.30\% & 68.90\% & 65.20\% \\

\rowcolor{dasGold}
& A-MEM
& 89.60\% & 85.60\% & 82.40\%
& 80.00\% & 72.80\% & 72.00\% \\

\rowcolor{dasGold}
& Memento
& 88.00\% & 85.60\% & 81.60\%
& 77.60\% & 75.20\% & 69.60\% \\

\rowcolor{dasGold}
\multirow{-4}{*}{\textbf{DASMark}}
& Evolving Harness
& 96.00\% & 90.30\% & 84.70\%
& 78.20\% & 66.90\% & 64.50\% \\


\rowcolor{dasGold}
& No-Mem
& \textbf{97.70\%} & \textbf{94.70\%}
& \textbf{92.90\%} & \textbf{82.60\%}
& \textbf{72.20\%} & \textbf{68.90\%} \\

\rowcolor{dasGold}
& A-MEM
& \textbf{97.60\%} & \textbf{95.20\%}
& \textbf{89.60\%} & \textbf{88.00\%}
& \textbf{80.80\%} & \textbf{74.40\%} \\

\rowcolor{dasGold}
& Memento
& \textbf{95.20\%} & \textbf{93.60\%}
& \textbf{89.60\%} & \textbf{87.20\%}
& \textbf{80.80\%} & \textbf{76.80\%} \\

\rowcolor{dasGold}
\multirow{-4}{*}{\textbf{DASMark-R}}
& Evolving Harness
& \textbf{96.80\%} & \textbf{95.20\%}
& \textbf{92.70\%} & \textbf{89.50\%}
& \textbf{80.60\%} & \textbf{78.20\%} \\

\bottomrule
\end{tabular}%
}

\vspace{4pt}

\textbf{(c) Synonym Substitution}\par\vspace{3pt}
\resizebox{\textwidth}{!}{%
\begin{tabular}{llcccccc}
\toprule

\multirow{2}{*}{\textbf{Method}}
& \multirow{2}{*}{\textbf{Backend}}
& \multicolumn{6}{c}{\textbf{Synonym Substitution Rate}} \\

\cmidrule(lr){3-8}

& & \textbf{10\%} & \textbf{20\%}
  & \textbf{30\%} & \textbf{40\%}
  & \textbf{50\%} & \textbf{60\%} \\

\midrule

& No-Mem
& 86.70\% & 58.30\% & 41.70\%
& 11.70\% & 3.30\% & 0.00\% \\

& A-MEM
& 75.80\% & 54.50\% & 27.30\%
& 12.10\% & 3.00\% & 0.00\% \\

& Memento
& 59.60\% & 38.60\% & 15.80\%
& 3.50\% & 0.00\% & 0.00\% \\

\multirow{-4}{*}{\textbf{AgentMark-F}}
& Evolving Harness
& 77.00\% & 45.90\% & 36.10\%
& 21.30\% & 21.30\% & 6.60\% \\

\midrule


& No-Mem
& 84.80\% & 83.30\% & 80.30\%
& 78.80\% & 69.70\% & 67.40\% \\

& A-MEM
& 84.80\% & 80.80\% & 79.20\%
& 72.80\% & 72.80\% & 67.20\% \\

& Memento
& 88.80\% & 89.60\% & 84.00\%
& 80.00\% & 76.00\% & 73.60\% \\

\multirow{-4}{*}{\textbf{KGW}}
& Evolving Harness
& 81.50\% & 78.20\% & 74.20\%
& 72.60\% & 66.90\% & 60.50\% \\

\midrule


\rowcolor{dasGold}
& No-Mem
& 90.20\% & 87.10\% & 84.80\%
& 75.00\% & 78.00\% & 76.50\% \\

\rowcolor{dasGold}
& A-MEM
& 91.20\% & 80.80\% & 76.00\%
& 74.40\% & 76.00\% & 73.60\% \\

\rowcolor{dasGold}
& Memento
& 93.60\% & 86.40\% & 82.40\%
& 76.80\% & 72.80\% & 69.60\% \\

\rowcolor{dasGold}
\multirow{-4}{*}{\textbf{DASMark}}
& Evolving Harness
& 92.70\% & 87.90\% & 86.30\%
& 80.60\% & 72.60\% & 73.40\% \\


\rowcolor{dasGold}
& No-Mem
& \textbf{95.50\%} & \textbf{90.80\%}
& \textbf{89.40\%} & \textbf{87.10\%}
& \textbf{82.60\%} & \textbf{82.60\%} \\

\rowcolor{dasGold}
& A-MEM
& \textbf{95.20\%} & \textbf{87.20\%}
& \textbf{84.00\%} & \textbf{81.60\%}
& \textbf{84.00\%} & \textbf{80.80\%} \\

\rowcolor{dasGold}
& Memento
& \textbf{95.20\%} & \textbf{91.20\%}
& \textbf{90.40\%} & \textbf{85.60\%}
& \textbf{84.80\%} & \textbf{76.80\%} \\

\rowcolor{dasGold}
\multirow{-4}{*}{\textbf{DASMark-R}}
& Evolving Harness
& \textbf{94.40\%} & \textbf{93.50\%}
& \textbf{89.50\%} & \textbf{88.70\%}
& \textbf{84.70\%} & \textbf{83.90\%} \\

\bottomrule
\end{tabular}%
}

\caption{Complete robustness results on Retail under (a) paraphrasing, (b) token deletion (10--60\%), and (c) synonym substitution (10--60\%) across four runtime backends. Values denote bit-decoding accuracy. DASMark-R uses a robustness-enhanced receiver. Bold values mark the best accuracy for each backend and attack setting (including ties).}
\label{tab:retail-robustness}
\label{tab:retail-semantic-rewriting}
\label{tab:retail-token-delete}
\label{tab:retail-synonym-replacement}
\end{table*}

\begin{table*}[p]
\centering
\footnotesize
\setlength{\tabcolsep}{4pt}
\renewcommand{\arraystretch}{1.0}
\setlength{\aboverulesep}{0pt}
\setlength{\belowrulesep}{0pt}

\textbf{(a) Paraphrasing (Semantic Rewriting)}\par\vspace{3pt}
\begin{tabular*}{\textwidth}{@{\extracolsep{\fill}}lcccc@{}}
\toprule

\textbf{Method}
& \textbf{No-Mem}
& \textbf{A-MEM}
& \textbf{Memento}
& \textbf{Evolving Harness} \\

\midrule

AM-F
& 76.40\%
& 70.00\%
& 70.50\%
& 80.30\% \\

KGW
& 76.80\%
& 77.60\%
& 80.30\%
& 82.40\% \\

\rowcolor{dasGold}
DASMark
& \textbf{88.40\%}
& 88.70\%
& 88.30\%
& 87.40\% \\

\rowcolor{dasGold}
DASMark-R
& 84.50\%
& \textbf{89.10\%}
& \textbf{90.40\%}
& \textbf{92.20\%} \\

\bottomrule
\end{tabular*}

\vspace{4pt}

\textbf{(b) Token Deletion}\par\vspace{3pt}
\resizebox{\textwidth}{!}{%
\begin{tabular}{llcccccc}
\toprule

\multirow{2}{*}{\textbf{Method}}
& \multirow{2}{*}{\textbf{Backend}}
& \multicolumn{6}{c}{\textbf{Token Deletion Rate}} \\

\cmidrule(lr){3-8}

& & \textbf{10\%} & \textbf{20\%}
  & \textbf{30\%} & \textbf{40\%}
  & \textbf{50\%} & \textbf{60\%} \\

\midrule


& No-Mem
& 15.90\% & 3.20\% & 0.50\%
& 0.50\% & 0.00\% & 0.00\% \\

& A-MEM
& 13.10\% & 1.40\% & 1.40\%
& 0.50\% & 0.00\% & 0.00\% \\

& Memento
& 12.90\% & 2.80\% & 0.50\%
& 0.00\% & 0.00\% & 0.00\% \\

\multirow{-4}{*}{\textbf{AgentMark-F}}
& Evolving Harness
& 10.90\% & 1.40\% & 0.00\%
& 0.00\% & 0.00\% & 0.00\% \\

\midrule


& No-Mem
& \textbf{89.70\%} & 81.20\% & 79.90\%
& 77.70\% & 70.10\% & 62.50\% \\

& A-MEM
& \textbf{93.50\%} & \textbf{87.90\%}
& \textbf{86.40\%} & 75.20\%
& 70.10\% & 62.60\% \\

& Memento
& \textbf{91.50\%} & 85.00\% & 80.80\%
& 75.60\% & 68.50\% & 59.20\% \\

\multirow{-4}{*}{\textbf{KGW}}
& Evolving Harness
& 93.40\% & 89.00\% & \textbf{84.90\%}
& 78.30\% & 71.30\% & 71.00\% \\

\midrule

\rowcolor{dasGold}
& No-Mem
& 89.20\% & \textbf{88.00\%} & 81.30\%
& 77.30\% & 70.90\% & 68.90\% \\

\rowcolor{dasGold}
& A-MEM
& 89.80\% & 85.30\% & 81.50\%
& 78.10\% & 75.50\% & 68.30\% \\

\rowcolor{dasGold}
& Memento
& 87.90\% & 84.00\% & 83.60\%
& 79.70\% & 74.70\% & \textbf{73.70\%} \\

\rowcolor{dasGold}
\multirow{-4}{*}{\textbf{DASMark}}
& Evolving Harness
& 90.10\% & 86.30\% & 81.20\%
& 79.90\% & 76.10\% & 74.10\% \\


\rowcolor{dasGold}
& No-Mem
& 86.90\% & 86.10\% & \textbf{85.70\%}
& \textbf{82.10\%} & \textbf{76.10\%}
& \textbf{74.10\%} \\

\rowcolor{dasGold}
& A-MEM
& 88.30\% & \textbf{87.90\%} & 85.30\%
& \textbf{80.00\%} & \textbf{81.10\%}
& \textbf{75.50\%} \\

\rowcolor{dasGold}
& Memento
& 91.10\% & \textbf{85.10\%}
& \textbf{85.10\%} & \textbf{80.10\%}
& \textbf{76.50\%} & 69.80\% \\

\rowcolor{dasGold}
\multirow{-4}{*}{\textbf{DASMark-R}}
& Evolving Harness
& \textbf{93.90\%} & \textbf{89.10\%}
& 84.00\% & \textbf{85.70\%}
& \textbf{82.60\%} & \textbf{78.20\%} \\

\bottomrule
\end{tabular}%
}

\vspace{4pt}

\textbf{(c) Synonym Substitution}\par\vspace{3pt}
\resizebox{\textwidth}{!}{%
\begin{tabular}{llcccccc}
\toprule

\multirow{2}{*}{\textbf{Method}}
& \multirow{2}{*}{\textbf{Backend}}
& \multicolumn{6}{c}{\textbf{Synonym Substitution Rate}} \\

\cmidrule(lr){3-8}

& & \textbf{10\%} & \textbf{20\%}
  & \textbf{30\%} & \textbf{40\%}
  & \textbf{50\%} & \textbf{60\%} \\

\midrule

& No-Mem
& 45.00\% & 18.60\% & 7.30\%
& 1.80\% & 1.40\% & 0.00\% \\

& A-MEM
& 39.90\% & 19.70\% & 6.60\%
& 2.80\% & 1.40\% & 0.50\% \\

& Memento
& 39.20\% & 14.30\% & 6.50\%
& 1.80\% & 0.00\% & 0.00\% \\

\multirow{-4}{*}{\textbf{AgentMark-F}}
& Evolving Harness
& 37.40\% & 11.60\% & 5.40\%
& 2.00\% & 1.40\% & 0.00\% \\

\midrule


& No-Mem
& 90.60\% & 87.90\% & 83.90\%
& 81.70\% & 75.60\% & 73.70\% \\

& A-MEM
& \textbf{93.00\%} & \textbf{91.10\%}
& 85.00\% & 83.60\% & 81.30\%
& 72.90\% \\

& Memento
& \textbf{94.40\%} & \textbf{89.70\%}
& 84.50\% & 79.30\% & 75.60\%
& 71.40\% \\

\multirow{-4}{*}{\textbf{KGW}}
& Evolving Harness
& \textbf{94.50\%} & 90.40\% & 89.30\%
& 87.50\% & 80.50\% & 79.00\% \\

\midrule

\rowcolor{dasGold}
& No-Mem
& 89.20\% & 85.70\% & 84.50\%
& 81.30\% & 82.10\% & 78.50\% \\

\rowcolor{dasGold}
& A-MEM
& 91.70\% & 90.60\% & \textbf{88.30\%}
& 84.90\% & 82.30\% & 80.80\% \\

\rowcolor{dasGold}
& Memento
& 90.70\% & 87.90\% & \textbf{87.20\%}
& \textbf{88.60\%} & \textbf{84.30\%}
& \textbf{82.20\%} \\

\rowcolor{dasGold}
\multirow{-4}{*}{\textbf{DASMark}}
& Evolving Harness
& 90.80\% & 90.80\% & 86.30\%
& 82.90\% & 82.30\% & 79.50\% \\


\rowcolor{dasGold}
& No-Mem
& \textbf{92.40\%} & \textbf{88.00\%}
& \textbf{87.60\%} & \textbf{86.10\%}
& \textbf{84.90\%} & \textbf{79.30\%} \\

\rowcolor{dasGold}
& A-MEM
& 90.20\% & 89.40\% & 87.50\%
& \textbf{86.80\%} & \textbf{86.80\%}
& \textbf{86.80\%} \\

\rowcolor{dasGold}
& Memento
& 89.00\% & 87.90\% & 86.80\%
& 84.00\% & 82.20\% & 80.10\% \\

\rowcolor{dasGold}
\multirow{-4}{*}{\textbf{DASMark-R}}
& Evolving Harness
& 93.90\% & \textbf{92.50\%}
& \textbf{89.80\%} & \textbf{89.10\%}
& \textbf{86.70\%} & \textbf{85.70\%} \\

\bottomrule
\end{tabular}%
}

\caption{Complete robustness results on CRMArena-Pro under (a) paraphrasing, (b) token deletion (10--60\%), and (c) synonym substitution (10--60\%) across four runtime backends. Values denote bit-decoding accuracy. DASMark-R uses a robustness-enhanced receiver. Bold values mark the best accuracy for each backend and attack setting (including ties).}
\label{tab:CRMArena-Pro-robustness}
\label{tab:CRMArena-Pro-semantic-rewriting}
\label{tab:CRMArena-Pro-token-delete}
\label{tab:CRMArena-Pro-synonym-replacement}
\end{table*}


\end{document}